\documentclass[journal]{vgtc}                % final (journal style)
\ifpdf%                                % if we use pdflatex
  \pdfoutput=1\relax                   % create PDFs from pdfLaTeX
  \usepackage{graphicx}                % allow us to embed graphics files
  \DeclareGraphicsExtensions{.pdf,.png,.jpg,.jpeg} % for pdflatex we expect .pdf, .png, or .jpg files
\else%                                 % else we use pure latex
  \usepackage{graphicx}                % allow us to embed graphics files
  \DeclareGraphicsExtensions{.eps}     % for pure latex we expect eps files
\fi%

\graphicspath{{figures/}{pictures/}{images/}{./}} % where to search for the images

\usepackage{microtype}                 % use micro-typography (slightly more compact, better to read)
\PassOptionsToPackage{warn}{textcomp}  % to address font issues with \textrightarrow
\usepackage{textcomp}                  % use better special symbols
\usepackage{mathptmx}                  % use matching math font
\usepackage{times}                     % we use Times as the main font
\usepackage{cite}                      % needed to automatically sort the references
\usepackage{tabu}                      % only used for the table example
\usepackage{booktabs}                  % only used for the table example
\usepackage{color}

\newcommand{\argmin}{\mathop{\rm arg~min}\limits}
\newcommand{\revise}[1]{\textcolor{black}{#1}}
\onlineid{1019}

\vgtccategory{Research}
\vgtcpapertype{algorithm/technique}

\title{Multifocal Stereoscopic Projection Mapping}

\author{Sorashi Kimura, Daisuke Iwai, \textit{Member, IEEE},\\ Parinya Punpongsanon, \textit{Member, IEEE}, and Kosuke Sato, \textit{Member, IEEE}}
\authorfooter{
\item
 Sorashi Kimura, Daisuke Iwai, Parinya Punpongsanon, and Kosuke Sato are with the Graduate School of Engineering Science, Osaka University. E-mail: E-mail: see https://www.sens.sys.es.osaka-u.ac.jp/.
\item
 Daisuke Iwai is with PRESTO, Japan Science and Technology Agency.
}

\shortauthortitle{Kimura \MakeLowercase{\textit{et al.}}: Multifocal Stereo Projection Mapping}
\abstract{
Stereoscopic projection mapping (PM) allows a user to see a three-dimensional (3D) computer-generated (CG) object floating over physical surfaces of arbitrary shapes around us using projected imagery.
However, the current stereoscopic PM technology only satisfies binocular cues and is not capable of providing correct focus cues, which causes a vergence--accommodation conflict (VAC).
Therefore, we propose a multifocal approach to mitigate VAC in stereoscopic PM.
Our primary technical contribution is to attach electrically focus-tunable lenses (ETLs) to active shutter glasses to control both vergence and accommodation.
Specifically, we apply fast and periodical focal sweeps to the ETLs, which causes the ``virtual image'' (as an optical term) of a scene observed through the ETLs to move back and forth during each sweep period.
A 3D CG object is projected from a synchronized high-speed projector only when the virtual image of the projected imagery is located at a desired distance.
This provides an observer with the correct focus cues required.
In this study, we solve three technical issues that are unique to stereoscopic PM: (1) The 3D CG object is displayed on non-planar and even moving surfaces; (2) the physical surfaces need to be shown without the focus modulation; (3) the shutter glasses additionally need to be synchronized with the ETLs and the projector.
We also develop a novel compensation technique to deal with the ``lens breathing'' artifact that varies the retinal size of the virtual image through focal length modulation.
Further, using a proof-of-concept prototype, we demonstrate that our technique can present the virtual image of a target 3D CG object at the correct depth.
Finally, we validate the advantage provided by our technique by comparing it with conventional stereoscopic PM using a user study on a depth-matching task.
} % end of abstract


\CCScatlist{ % not used in journal version
 \CCScat{K.6.1}{Management of Computing and Information Systems}%
{Project and People Management}{Life Cycle};
 \CCScat{K.7.m}{The Computing Profession}{Miscellaneous}{Ethics}
}

\teaser{
  \centering
  \includegraphics[width=\linewidth]{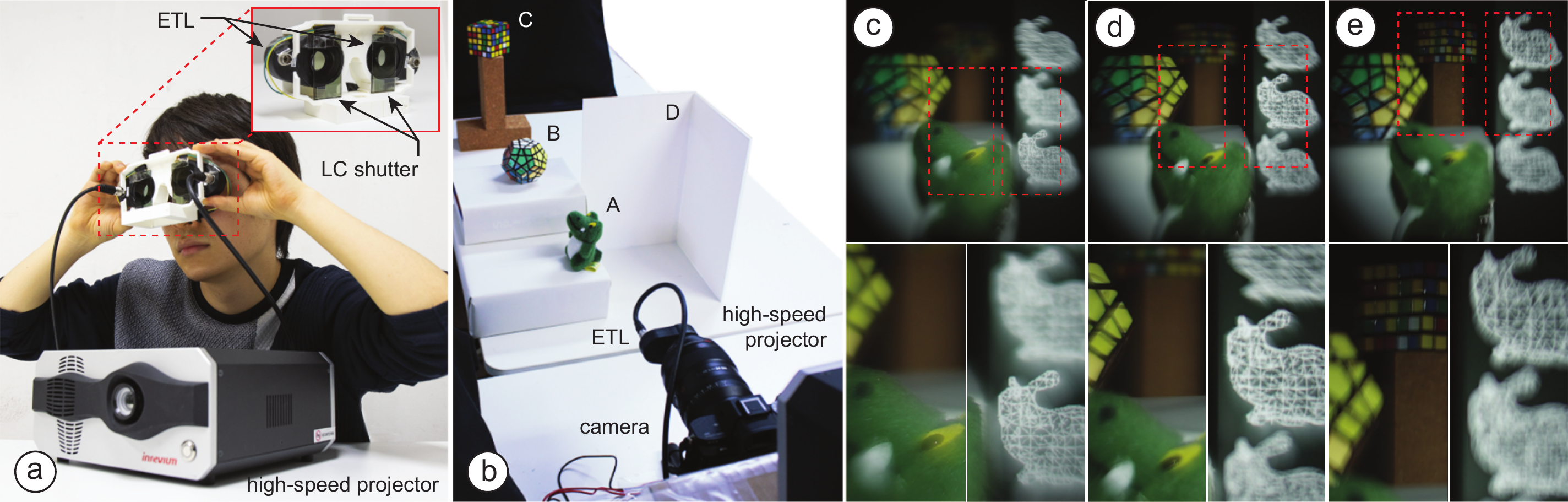}
  \caption{The proposed multifocal stereoscopic projection mapping technique mitigates the vergence--accommodation conflict by placing the virtual image (as an optical term) at the desired distance from an observer using focal-sweep active shutter glasses and a synchronized high-speed projector. (a) Prototype system; (b) experimental setup to show the multifocal appearance of Stanford bunnies onto a flat surface (D) whose virtual images were placed, from near to far, at points at which three physical objects (A--C) were placed; (c--e) captured scene observed through a focal sweep lens. The bunnies, as well as the physical objects, appeared focused from the bottom to the top according to the focus distance of the camera lens at (c) near, (d) middle, and (e) far.}
	\label{fig:teaser}
}

\vgtcinsertpkg

\begin{document}


\firstsection{Introduction}

\maketitle

%% \section{Introduction} %for journal use above \firstsection{..} instead

% ステレオPMの応用可能性
Stereoscopic projection mapping (PM) enables a user to see a three-dimensional (3D) computer-generated (CG) object floating over the physical surfaces of arbitrary shapes around them using projected imagery.
This technique can also be used to visually deform the surfaces.
These illusory effects are achieved by tracking an observer's viewpoint, rendering the perspectively correct images with proper disparity for each eye, and projecting the two images in a time-sequential manner in each frame.
The projected images are observed through active-shutter glasses consisting of liquid crystal (LC) shutters, which prevent interference between the two images for both eyes.
There are several technical advantages of using stereoscopic PM over other types of augmented reality (AR) displays such as optical see-through (OST) and video see-through (VST) head-mounted displays (HMD).
First, the field-of-view (FOV) can be made as wide as possible by increasing the number of projectors to cover the entire environment.
Second, the active-shutter glasses used are normally much lighter, and, thus, their physical burden is less than HMDs.
Third, multiple users can share the same AR experience if their viewpoints are sufficiently close to each other.
Thanks to these advantages, researchers have found stereoscopic PM to be suitable for a wide range of applications, including but not limited to museum guides \cite{SCHMIDT20191}, architecture planning \cite{1544657}, product design \cite{8797923}, medical training \cite{kijima07}, shape-changing interfaces~\cite{10.1145/2858036.2858457}, and teleconferencing \cite{10.1145/2207676.2207704}.

% VAC問題と、これまでの大まかな解法。これまではディスプレイ単体しか対象ではなかった。スクリーンまでの距離がそもそも一定ではなく、常に変化する状況での手法はない
In many of these applications, it is important to create a faithful impression of the 3D structure of the projected object or scene.
However, like most commercial virtual reality (VR) and AR displays, current stereoscopic PM only satisfies binocular cues and is not capable of providing the correct focus cues, which causes a problem called the vergence--accommodation conflict (VAC).
%Stereoscopic PM can provide correct binocular cues by the well-established computational technique of projection images \cite{10.5555/1088894}.
Vergence refers to the simultaneous movement of both eyes in opposite directions to obtain single binocular vision, while accommodation refers to the process by which each eye changes its optical power to focus on an object as its distance varies.
The uncoupling of the vergence and accommodation causes significant discomfort, fatigue, and a distorted 3D perception for the observer \cite{Hoffman2008}.
While many approaches have been proposed to mitigate the VAC in VR/AR displays (see Sect.~\ref{sec:relatedwork}), to the best of our knowledge, no technical solution has yet been presented for the VAC in stereoscopic PM.

% 提案する
We propose a multifocal approach to improve the mismatch between vergence and accommodation in stereoscopic PM.
Our primary technical contribution is to attach electrically focus-tunable lenses (ETL) to active-shutter glasses to control both vergence and accommodation.
Specifically, we apply fast and periodical focal sweeps to the ETLs, which causes the ``virtual image'' (as an optical term) of every part of the real scene seen through the ETLs to move back and forth during each sweep period.
In each frame, we first render the perspectively correct images of a 3D CG object for both eyes.
%using the observer's viewpoint, the shapes of projection surfaces, and the relative pose of a projector to the surfaces.
Then, we project them from a synchronized high-speed projector at the exact moment that the virtual image of the projected imagery on a real surface is located at a desired distance from the ETLs.
Thus, each part of the surface is projected for a short period of time during each sweep period.
The frequency of the focal sweep is higher than the critical flicker fusion frequency (CFF) so that the observer perceives the time integral of the scene appearances during the sweep period.
%When the focusing distance is modulated from near to far, rendered images for one of the eyes are projected while the shutter glass of the other eye is closed.
%When it is modulated back from far to near, rendered images for the other eye are projected.

This paper introduces the technical details of our computational algorithms, which are jointly designed with the optics and hardware.
Particularly, we solve three technical issues that are unique to stereoscopic PM: (1) The 3D CG object is displayed on non-planar and even moving surfaces; (2) the physical surfaces that are also seen through the ETLs need to be shown without the focus modulation; (3) the LC shutter glasses also need to be synchronized with the ETLs and the projector to provide the correct binocular cues as well as the focus cues.
On the other hand, in other AR/VR displays, (1) the 3D CG object is displayed on a planar and fixed display panel, (2) observers do not see a real scene through ETLs, and (3) shutter glasses are not used.
We also develop a novel compensation technique to address the ``lens breathing'' artifact, which varies the retinal size of the virtual image through focal length modulation.
Using a proof-of-concept prototype, we demonstrate that our technique can present the virtual image of a target 3D CG object with the perspectively correct appearance at the correct depth.
% apparent origins of ray divergences.
Finally, we validate the advantage provided by our technique with regard to the mitigation of the VAC by comparing it with a conventional stereoscopic PM using a user study on a depth-matching task.

To summarize, our primary contributions are that we
\begin{itemize}
    \item introduce a multifocal stereoscopic PM technique that mitigates the VAC by combining fast and periodical focal sweeps with active-shutter glasses and synchronizing high-speed projection,
    \item jointly design optics, hardware, and computational algorithms to present a 3D CG object with correct binocular and focus cues, solve three technical issues unique to stereoscopic PM, and propose a general computational solution for the lens breathing artifacts,
    \item implement a proof-of-concept prototype that displays an object's virtual image that is perspectively correct and located at a desired distance from the observer, and
    \item demonstrate the VAC mitigation achieved by the proposed system using a user study on a depth-matching task.
\end{itemize}

%\begin{figure}[t]
%  \centering
%  \includegraphics[width=0.98\linewidth]{figure/figure-proposed-system-1}
%  \caption{(draft) Proposed method -- not link to the content yet.}
%  \label{fig:proposed-method}
%\end{figure}

\section{Related Work}
\label{sec:relatedwork}

%This paper presents a technical solution for VAC in stereoscopic PM.
%We introduce previous works of stereoscopic PM and VAC mitigation methods for other types of VR/AR displays, and state our contributions compared to them.

\subsection{Stereoscopic PM}\label{subsec:rel_pm}

Since the pioneering work conducted on rendering perspectively correct images on non-planar physical surfaces using projected imagery~\cite{10.1145/280814.280861}, much research has been carried out on stereoscopic PM.
%A large body of research applied the rendering technique, which is so-called ``two-pass rendering''~\cite{10.1145/280814.280861}.
%It consists of the following off-screen and on-screen rendering passes conducted in a virtual space where the positions of an observer and a projector as well as the shape of a projection surface are incorporated.
%First, off-screen rendering is performed to generate the desired appearance from the observer's viewpoint.
%The rendered image is then projected onto the surface from the observer's viewpoint in the virtual space.
%Then, on-screen rendering is performed, in which the newly textured surface is captured from the projector's viewpoint.
%The captured image is finally projected in the physical space.
%
Due to several advantages, such as the wider FOV and fewer physical burdens than other types of VR/AR displays, stereoscopic PM has been applied in various fields.
Telepresence has been considered a particularly suitable application of stereoscopic PM.
The 3D shape and appearance of a distant person are captured using an RGB-D camera and displayed on local physical surfaces using projected imagery to make it appear as the distant person exists in the same room~\cite{10.1145/2818048.2819965,10.1145/2642918.2647402} or behind a wall~\cite{10.1145/280814.280861,10.1145/2207676.2207704}.
Stereoscopic PM has also been applied in architecture planning \cite{1544657,egsh.20171001}, where new structures of a building are displayed on a wall of the building.
In the car industry, it was found that stereoscopic PM has the potential to assist in interior designing such that designers can examine the shape of a dashboard without needing to fabricate physical mock-ups \cite{8797923}.
Stereoscopic PM has also been applied in medical training~\cite{kijima07}, where human organs inside a body are projected on the body.
By focusing on expanding the application fields, previous works have not sufficiently considered the VAC, which, thus, remains an unsolvable and challenging technical issue in stereoscopic PM.

Various types of human perceptual properties in stereoscopic PM have been investigated.
For example, Okutani et al. found that stereoscopic PM induces stereoscopic capture where the depth observed during stereoscopic PM is automatically attributed to the texture of a physical surface even when the texture conveys no disparity information~\cite{8491263}.
Other researchers found that pseudo-haptic illusions occur in stereoscopic PM, where a user can feel bumps on a physically flat surface on which apparent bumps are overlaid\cite{8417321}.
Depth perception in stereoscopic PM has been more intensively analyzed than the other properties.
Certain researchers tested the depth estimation ability of human participants given the binocular disparity in stereoscopic PM~\cite{egve.20171354,10.1145/3359996.3364245}, while others did it with additional apparent cues such as color and texture~\cite{Schmidt2020}.
However, due to the lack of a stereoscopic PM system that satisfies both binocular and focus cues, how the depth estimation accuracy is affected when the VAC is mitigated has not yet been investigated.

%projection location optimization \cite{10.1145/3173574.3173843}

\subsection{Mitigating the VAC}\label{subsec:rw_vac}

% VACは問題なので、解決する方法がたくさん提案されている
% VRディスプレイにおいて、ソフトウェアベースのDOFシミュレーションでは十分では無いことが示唆されており、ハードウェアによる解決方法がたくさん提案されてきた。
% focusとfixationをマッチする技術として、varifocal, multifocal, light field, holographicなどが提案されている
% ARでも同様に、varifocal, multifocal, light field, holographicが提案されている。
% 一方、これらはいずれもconventional stereoscopic displayかnear eye display (HMD)を対象にしたものであり、プロジェクションマッピングを対象としたものをto the best of our knowledge知らない。

% 我々は、ETLをメガネとして用い、プロジェクタからの照明タイミングを同期させることで、実空間のfocus cueを操作することを実現した
% inspired by this work　本研究では、実空間ではなくそこに投影されるバーチャル像のfocus cueを操作することを目指す
% 立体投影と組み合わせるのは新しい。
% lens breathing問題を、computationalに解決する（手動ではない）ことも新しい。

In VR and AR applications, CG images need to create a faithful impression of the 3D structure of the depicted object or scene.
Unfortunately, commercially available 3D displays often yield perceptual distortions in the 3D structure due to the mismatch between binocular and focus cues.
The VAC causes viewer fatigue and discomfort as well as distortions in the perceived depth~\cite{WANN19952731,Hoffman2008}.
Much research has been carried out in an attempt to solve the VAC issue in both VR and AR displays, which include varifocal displays (VR~\cite{Padmanaban2183,10.1145/2858036.2858140,Johnson:16,10.1145/3072959.3073622} and AR~\cite{10.1145/3272127.3275069,10.1145/3130800.3130892,8458263,8794584,Chen:15,7829412}), multifocal displays (VR~\cite{10.1145/3272127.3275015,10.1145/2766909,10.1145/1015706.1015804,10.1145/3072959.3073590,10.1145/3386569.3392424,6712903,Love:09,10.1145/3130800.3130846,Rolland:00} and AR~\cite{8456852,4637321,9284672,Llull:15,Chen:19,https://doi.org/10.1002/jsid.739,Basak:19}
), light field displays (VR~\cite{10.1145/2508363.2508366,10.1145/2766922,10.1145/2601097.2601144} and AR~\cite{6671761,10.1145/2601097.2601141}), holographic displays (VR~\cite{10.1145/3414685.3417846,10.1145/3414685.3417802,10.1145/3355089.3356539,10.1145/3386569.3392416} and AR~\cite{10.1145/3072959.3073624,10.1145/3272127.3275069}
), and \revise{other display techniques, such as those that use color information~\cite{Aksit:14,10.1145/3130800.3130815}}.
Unfortunately, despite the substantial effort that has been put into solving the VAC for stereoscopic flat-panel displays and HMDs, researchers have not focused on the issue for stereoscopic PM.
Previous works closely related to stereoscopic PM are associated with light field \cite{10.1145/2601097.2601144} and tomographic \cite{10.1145/3355089.3356577} projection systems.
These technologies have been developed for conventional flat projection screens and, thus, are not directly applied to PM, where the projection target includes potentially non-planar and even moving surfaces.
Furthermore, these works required customized, special projection devices, which potentially hinder developers from using them in the actual fields of application.

Among the four display technologies mentioned above, light field and holographic displays are not suitable for stereoscopic PM.
Since it is not assumed to install tailored optical elements in the environment in general PM applications, light field and holographic displays---which, for instance, need to embed an angle-expanding material~\cite{10.1145/2601097.2601144} and a phase modulator~\cite{10.1145/3072959.3073624} onto physical surfaces---are not generally applicable for stereoscopic PM.
Varifocal and multifocal displays basically share the same optical display architecture---both apply focus-tunable near-eye optics for displaying 3D CG objects with correct focus cues.
%A beam splitter is applied to allow a user to observe a real scene without any focus modulations.
%because both need to show images at different focal planes.
The major difference between the two displays is that varifocal displays actively track the accommodation of the eye in real-time and provide a correct image at the interested depth, while multifocal displays passively offer multiple depths to simulate a 3D scene without considering where the eye focuses~\cite{Zhan2020}.
%Because PM assumes that a user is freely moving in the environment, it is required to accurately measure the position of the user's viewpoint to precisely track the user's eye gaze.
%Even a slight error in the calibration and position estimation may cause critical artifacts in the displayed result.
\revise{In stereoscopic PM, the real scene is also seen through the focus-tunable near-eye optics because a 3D CG object is projected directly onto a surface in the scene. Therefore, when a varifocal approach is employed, a user needs to observe the real scene with a focal setting that is optimized for the projected 3D CG object. This focus modulation can possibly create an unnatural view of the real scene.}
%\revise{Thus, the varifocal approach is also not suitable in stereoscopic PM.}
In addition, the projection surface can be non-planar.
In this case, the projected 3D CG object is not positioned at the same distance from the observer, and, thus, a spatially-varying focal control is required to provide the correct focus cues.
However, a varifocal approach assumes a spatially uniform operation.
Furthermore, it is not preferable to add an eye-tracking sensor system to the active-shutter glasses because it degrades the advantage of stereoscopic PM (i.e., it only requires a lightweight eye wear) over the other VR/AR displays.
Therefore, we adopt the multifocal approach in this research to mitigate the VAC in stereoscopic PM.

User studies have revealed that the observer's depth estimation performance can be significantly improved by mitigating the VAC in both VR~\cite{Padmanaban2183,10.1145/2858036.2858140,Johnson:16,10.1145/3072959.3073622} and AR~\cite{7164348,8462792} displays.
Therefore, this improvement could happen in stereoscopic PM as well.
However, as mentioned in Sect.~\ref{subsec:rel_pm}, there is no experimental evidence of this due to the lack of a PM system satisfying both binocular and focus cues.

\subsection{Our contributions}

This paper presents the first stereoscopic PM system that mitigates the VAC by applying a multifocal approach.
On top of the conventional stereoscopic PM framework, we apply additional optics (ETLs) and computational algorithms to achieve this goal.
In particular, we describe how to solve the technical issues that are unique to stereoscopic PM, one of which is that the display surface is potentially non-planar and moving.
An advantage of our technique is that it does not need customized, special projection devices and works with off-the-shelf projectors.
%We also propose a general computational solution for the lens breathing artifacts, which can be used in the other varifocal and multifocal AR/VR displays applying focus tunable lenses.
We also conduct the first user study to investigate how the VAC mitigation improves the depth estimation accuracy in stereoscopic PM using a prototype system.

\section{Method}

%This section describes the details of our technique.
%First, we explain how to determine the optical power of an ETL to locate the virtual image of projected imagery onto a physical surface at a desired distance from the ETL.
%Second, our synchronization strategy of the projector, ETLs, and shutter glasses is introduced.
%We then propose a lens breathing compensation technique, and finally explain the whole rendering pipeline.

\subsection{Virtual image of projected imagery seen through ETL}\label{subsec:method_vi}

\begin{figure}[t]
  \centering
  \includegraphics[width=0.98\linewidth]{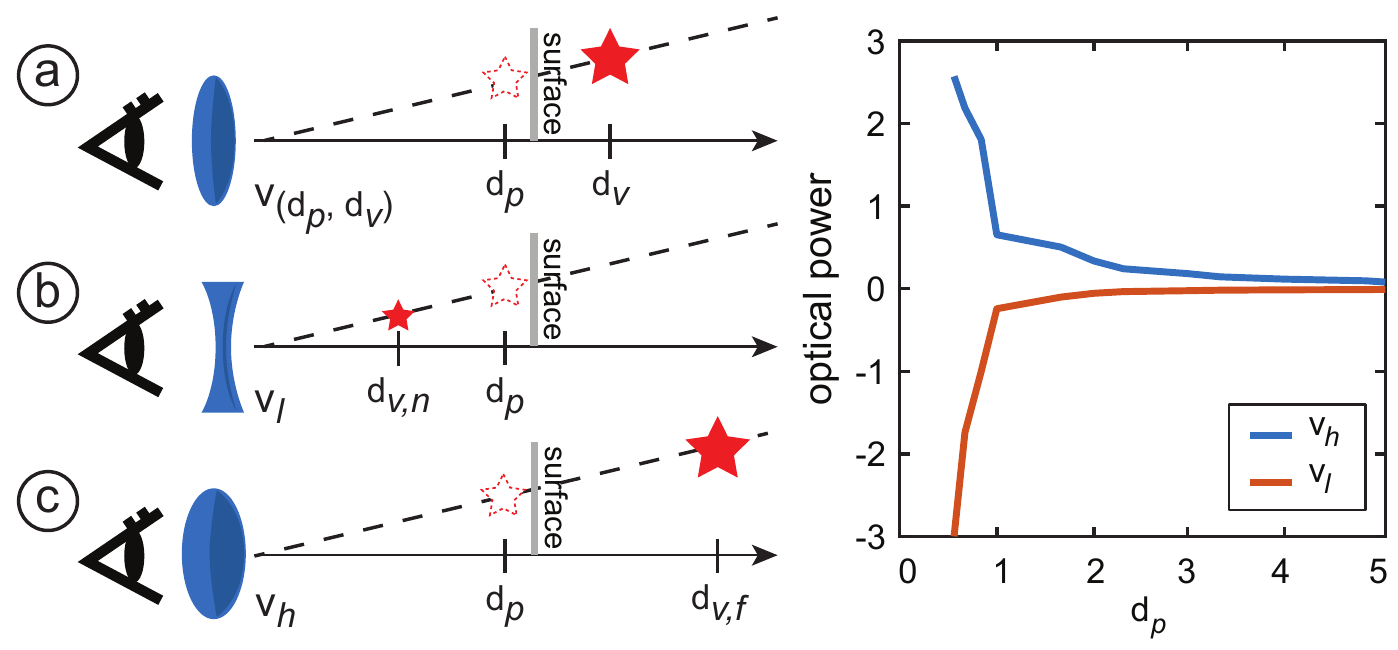}
  \caption{(Left) The virtual image (solid star) of a projected object (dashed star) on a physical surface observed through an ETL. (a) The virtual image is located at $d_v$ from the ETL when the projected image is at $d_p$, and the ETL's optical power is $v(d_p,d_v)$. (b)(c) When the optical power is periodically swept between $v_l$ and $v_h$, the virtual image also periodically moves between the near end $d_{v,n}$ and the far end $d_{v,f}$. (Right) The required sweep range of an ETL's optical power when $d_p$ is changed under the condition that $d_{v,n}$ and $d_{v,f}$ are fixed at $\frac{1}{6}$ m and $\frac{1}{0.4}$ m, respectively.}
  \label{fig:sweep-condition-concept}
\end{figure}

\revise{We attach an ETL each to the left and right of LC shutter of the active-shutter glasses.}
As a result, a user can observe the virtual image of the projected imagery on a physical surface through the ETL.
Suppose that the distance of the surface from the ETL is $d_p$ (Fig. \ref{fig:sweep-condition-concept}a).
Using the thin lens equation, we can determine the optical power $v(d_p,d_v)$ to locate the virtual image of a 3D CG object projected on the surface at a desired distance $d_v$ from the ETL as follows:
\begin{equation}\label{eq:thin_lens}
    v(d_p,d_v)=\frac{1}{d_p}-\frac{1}{d_v}.
\end{equation}
The optical power is the reciprocal of the focal length (m) and is generally represented in diopter (D).
In this paper, we do not fix $d_p$ because the surface can be non-planar and can move in stereoscopic PM.
Therefore, we place the virtual image of a projected 3D CG object at a desired distance by adjusting the optical power $v(d_p,d_v)$ according to both $d_p$ and $d_v$.

%By solving Eq. \ref{eq:thin_lens}, we can determine the optical power $v(d_p,d_v)$ by which the virtual image of a 3D CG object projected onto a surface at $d_p$ can be located at a desired distance $d_v$ as:
%
%\begin{equation}\label{eq:thin_lens2}
%    v(d_p,d_v)=\frac{d_v-d_p}{d_pd_v}
%\end{equation}
%

We apply a fast focal sweep to the ETL, where its optical power is periodically modulated between $v_l$ and $v_h$ ($v_l<v_h$) at $f$ Hz such that $f$ is higher than the CFF.
When the physical surface is lit by normal room lighting, the virtual image of the surface periodically moves between the near end (when the optical power is $v_l$) and the far end (when it is $v_h$) along the ETL's optical axis.
In our system, we turn off the room light and project the desired 3D CG object onto the surface when the optical power of the ETL becomes $v(d_p,d_v)$ for each focal sweep period.
We can locate the virtual image at the desired distance $d_v$ under the condition that the optical power is within the range of $v_l\le v(d_p,d_v)\le v_h$.

%When the desired distance of a 3D CG object $D'$ is within the range, we can represent it at the correct distance by projecting the perspectively-correct image when $f$ becomes 

%\begin{figure}[t]
%  \centering
%  \includegraphics[width=0.90\linewidth]{figure/fig-3-7-1}
%  \caption{The required sweep range of an ETL's optical power, when $d_p$ is changed under the condition that $d_{v,n}$ and $d_{v,f}$ are fixed at $\frac{1}{6}$ m and $d_f=\frac{1}{0.4}$ m, respectively.}
%  \label{fig:sweep-data}
%\end{figure}

% 近点と遠点
Given the desired location range of the virtual image of a projected 3D CG object and the position of a projection surface, the sweep range required to place the virtual image at the desired location is determined.
Suppose that the distance from the near end of the virtual image to the surface is $d_{v,n}$ and that from the far end to the surface is $d_{v,f}$ (Figs. \ref{fig:sweep-condition-concept}b and \ref{fig:sweep-condition-concept}c), the relationship between $v_l$ and $d_{v,n}$ and that between $v_h$ and $d_{v,f}$ can be formulated as follows according to the thin lens equation:
\begin{equation}\label{eq:sweep-range}
    v_l=\frac{1}{d_p}-\frac{1}{d_p-d_{v,n}},\ v_h=\frac{1}{d_p}-\frac{1}{d_p+d_{v,f}}.
\end{equation}
The right-hand side graph in Fig.~\ref{fig:sweep-condition-concept} shows the required sweep range computed by Eq.~\ref{eq:sweep-range} when $d_p$ is changed from $\frac{1}{3}$ m to $\frac{1}{0.2}$ m under the condition that $d_{v,n}$ and $d_{v,f}$ are fixed at $\frac{1}{6}$ m and $\frac{1}{0.4}$ m, respectively.
The graph indicates that the required sweep range becomes larger when the distance between the ETL and the surface is shorter.
Therefore, we determine the sweep range from the desired location range of the virtual image of a 3D CG object projected onto the nearest surface from the ETL.

\begin{figure}[t]
  \centering
  \includegraphics[width=0.98\linewidth]{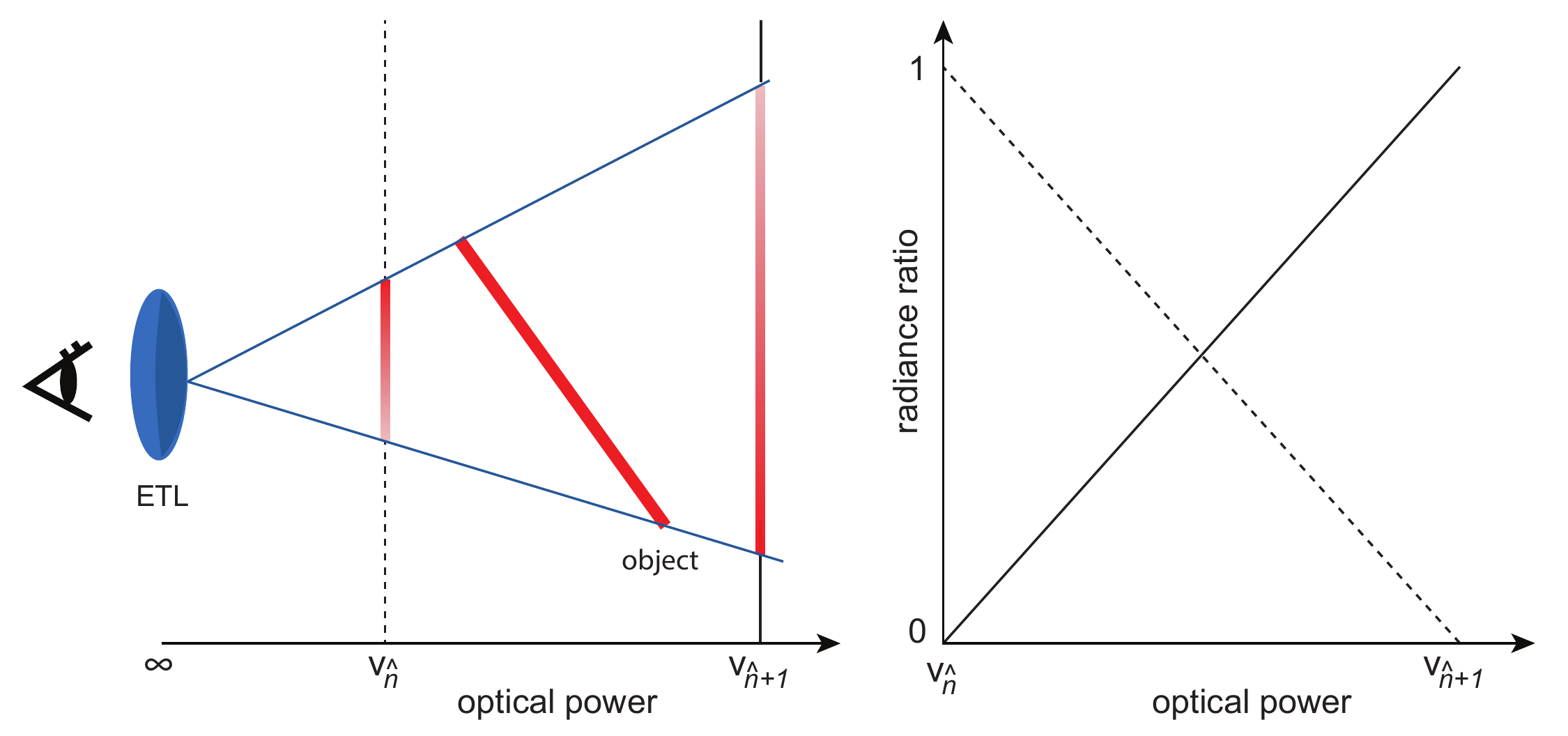}
  \caption{Depth filtering technique. (Left) A depicted example showing the radiance distribution between the two nearest projector frames and (right) radiance distribution in linear proportion with distance measured in diopters.}
  \label{fig:system-requirement}
\end{figure}

%近くのfocal planeに分散する
In an actual setup, the projector can display only a limited number of images ($N$) during a sweep period.
Therefore, it is not always possible to project a 3D CG object when the optical power of the ETL is exactly $v(d_p,d_v)$.
To overcome this constraint, we sample $N'$ ($\le N$) optical powers $v_n$ ($n=1,\cdots,N'$) within the range between $v_l$ and $v_h$ such that the projector can display an image at each $v_n$.
Then, a 3D CG object is projected at the $\hat{n}$-th optical power in each sweep period when $v_{\hat{n}}$ is the closest to $v(d_p,d_v)$; thus,
\begin{equation}\label{eq:divide}
    \hat{n}=\argmin_n |v_n-v(d_p,d_v)|.
\end{equation}
%
%depth filtering
The sampling of the optical power causes a discontinuity in the virtual image of a displayed 3D CG object when the optical power $v(d_p,d_v)$ is spatially varying and consequently spans two sampled optical powers (e.g., $v_{\hat{n}}$ and $v_{\hat{n}+1}$) (Fig. \ref{fig:system-requirement}).
To alleviate this discontinuity, we apply the ``depth filtering'' technique introduced in previous work~\cite{10.1145/1015706.1015804}.
Depth filtering distributes the radiance between the two nearest projector frames in linear proportion, with distance measured in diopters.
Suppose $r_{\hat{n}}$ and $r_{\hat{n}+1}$ are the distributed radiance of the original radiance $r$ to the $\hat{n}$-th and $(\hat{n}+1)$-th projection images, respectively; they would be computed as follows:
\begin{equation}\label{eq:depthf}
    r_{\hat{n}}=r(1-\frac{v(d_p,d_v)-v_{\hat{n}}}{v_{\hat{n}+1}-v_{\hat{n}}}),\ r_{\hat{n}+1}=r-r_{\hat{n}}.
\end{equation}
A previous user study showed that depth filtering could provide continuous focus cues \cite{MacKenzie2010}.
Another work recommended that the sampling interval of the optical power should be less than 0.6 D to provide perceptually continuous focus cues~\cite{10.1117/12.908883}.
We follow this guideline when we build our prototype.

\subsection{Synchronization}\label{subsec:method_sync}

%ガントチャート的な説明(ETL、シャッタグラス、投影タイミングの関係)
To display the virtual image of a 3D CG object at the desired location, we need to synchronize the projector, ETLs, and shutter glasses.
We determine our synchronization strategy while considering the following temporal response properties of the devices:
\begin{description}
    \item[Projector:] We employ a state-of-the-art off-the-shelf high-speed projector~\cite{watanabe2015high}. It projects an image at $\delta t_p$ ms after a trigger signal is sent from the PC. It is also known that the delay does not vary over time. Therefore, once we measure $\delta t_p$ in the calibration phase, we can project images at the desired timings by sending the triggers at $\delta t_p$ prior to the projection. There is also a delay when transferring an image from the PC to the projector ($\sim$3 ms). We apply a multi-threading pipeline to perform the projection and transfer the images in parallel, and, thus, the delay of the latter process does not have to be considered during synchronization.
    \item[ETLs:] We use polymer-based ETLs in which the lens consists of an optical fluid sealed off by an elastic polymer membrane. An electromagnetic actuator ring exerts pressure on the outer zone of the lens, which changes its curvature, i.e., the optical power. Polymer-based ETLs can achieve a faster focal change than other ETL types while maintaining a relatively large aperture size. As previous focal sweep systems~\cite{8456852,10.1145/3272127.3275015,10.1145/3072959.3073594,10.1145/3355089.3356577} have shown, the optical power of an ETL can be smoothly modulated when a sinusoidal or triangular wave is applied as the electrical current. Therefore, we apply a sinusoidal wave as the input in our system. It was also experimentally validated that the temporal response of an ETL is a linear time-invariant (LTI) system~\cite{Iwai2019}. Therefore, when we supply a sinusoidal wave $i[\phi]$ to an ETL, the output optical power also becomes a periodical wave $v[\phi]$, where $\phi$ is the phase of the input wave signal ($0\le\phi<2\pi$). Once we measure one cycle of $v[\phi]$ in the calibration phase, we can estimate the optical power at any timing required.
    \item[Shutter glasses:] We use LC-based active-shutter glasses. The LC is transparent (open) in the initial state and becomes opaque (closed) when a voltage is applied. There are two delays to be considered. First, a delay $\delta t_{o2c}$ occurs between the moment we apply the voltage to change the LC state from open to closed till the moment the LC shutter actually closes. Second, there is a longer delay $\delta t_{c2o}$ that occurs when it shifts from the closed to the open states as well. Therefore, we measure these delays during the calibration phase and use them for synchronization.
\end{description}

\begin{figure}[t]
  \centering
  \includegraphics[width=0.98\linewidth]{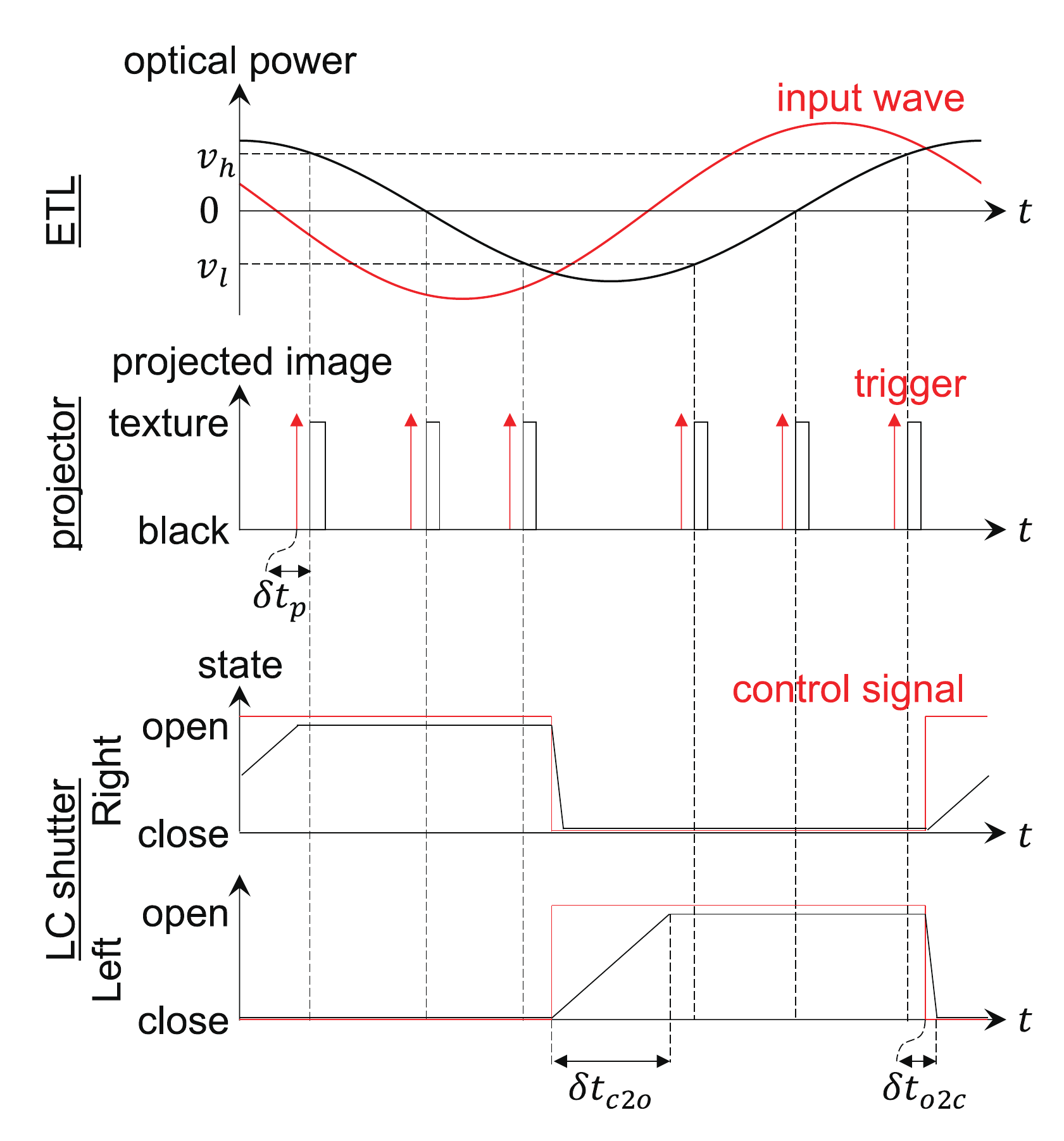}
  \caption{Timing chart of the proposed synchronization strategy. In this example, the number of the sampled ETL's optical power is three (i.e., $N'=3$), at which the high-speed projector projects images.}
  \label{fig:timing-chart}
\end{figure}

Figure~\ref{fig:timing-chart} shows the timing chart of our synchronization strategy.
First, the ETLs for the left and right eyes are modulated by supplying them with the same sinusoidal waves $i[\phi]$ of frequency $f$ Hz.
We divide the optical power waveform into two parts in each period and use the upward curve (from $v_l$ to $v_h$) for the left eye and the downward curve (from $v_h$ to $v_l$) for the right eye.
% プロジェクタ投影タイミング
The projector projects images when the optical power $v[\phi]$ of each ETL becomes $v_n$.
To make this happen, we send a trigger signal to the projector when the phase of the ETL's input signal is $2\pi f\delta t_p$ prior to the target phase.
% LCのタイミング
We open the LC of the left eye and close the LC of the right when the ETLs' optical powers form the upward curves.
In the same way, we open the LC of the right eye and close the LC of the left when the optical powers form the downward curves.
We control the LC shutters such that the optical powers become equal to the maximum and minimum values during the closing/opening transition time of the LC shutters because the time derivative of the modulated optical power of each ETL becomes smallest at these timings.

%実物体照明 まず、なぜETLをゼロジオプトリを取らせるか、というところから導入する
Stereoscopic PM applications generally project a white color onto physical surfaces to make them visible in addition to the projected 3D CG object to enable AR experiences.
In our system, the physical surfaces under white illumination are also observed through the ETL.
To avoid a situation in which the locations of the virtual images of these surfaces vary according to the focal sweep, we illuminate them when the optical powers of the ETLs are zero (i.e., $v[\phi]=0$).
This allows us to present the physical surfaces without any focus modulations.

\subsection{Lens breathing compensation}\label{subsec:method_comp}

The virtual image seen through the ETL is scaled using the change in its optical power because an observer's eye is not co-located with the ETL but, instead, placed behind it.
When a projected 3D CG object is divided into multiple parts that are observed through the ETL with different optical powers, the lens breathing phenomenon causes an overlap or gap between their virtual images on the retina.
This artifact has been noticed in previous works~\cite{10.1145/3072959.3073622,10.1145/3355089.3356577,10.1145/2766909}, in which it was solved by manually adjusting the size of displayed images.
In this paper, we propose a simple yet general computational solution for this issue.

% lens breathing現象を図解。ギャップとオーバーラップ
% 補償の方法も図解
\begin{figure*}[t]
  \centering
  \includegraphics[width=0.98\linewidth]{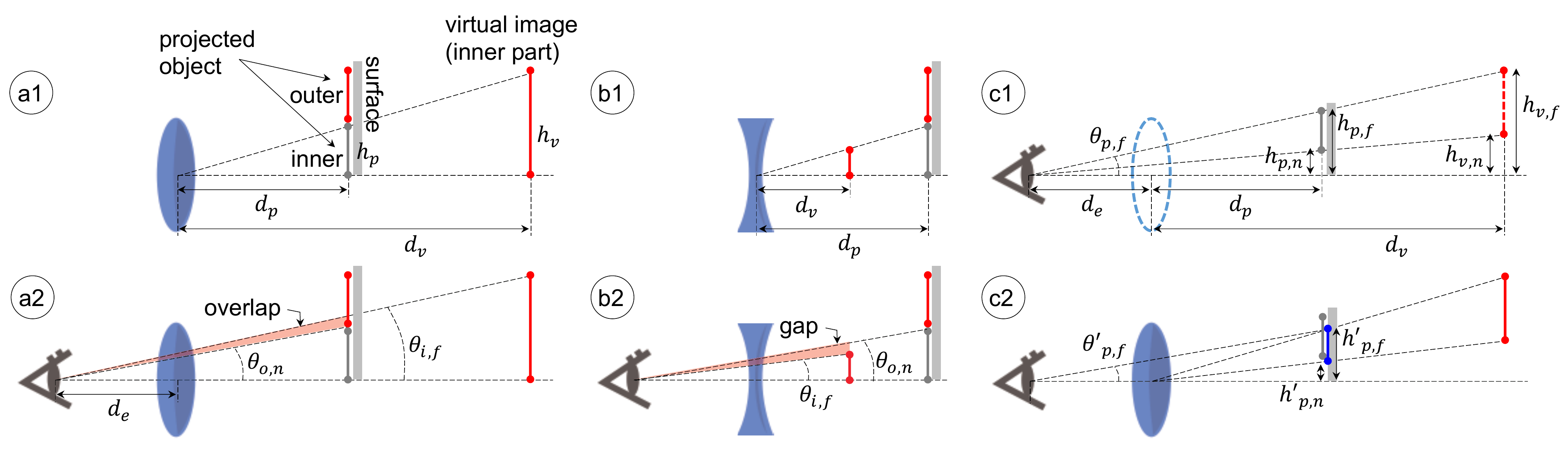}
  \caption{Lens breathing artifact and compensation. When the inner part of a projected object is observed with a (a) positive or (b) negative optical power, an (a) overlap or (b) gap occurs between its virtual image and that of the outer part observed with the optical power of zero. Note that \textbf{\textcolor{red}{---}} indicates an observed real image and the virtual image of \textbf{\textcolor[rgb]{0.5,0.5,0.5}{---}}. (c) The proposed compensation technique resizes the projected object to \textbf{\textcolor{blue}{---}}.}
  \label{fig:lens-breathing-compensation}
\end{figure*}

Without loss of generality, we suppose simple situations where a 3D CG object is projected onto a flat surface that is located $d_p$ away from an ETL (Figs.~\ref{fig:lens-breathing-compensation}(a1) and \ref{fig:lens-breathing-compensation}(b1)).
The object is divided into two parts that are observed with different optical powers.
When the optical power becomes equal to a certain non-zero value, the inner part of the object with respect to the ETL's optical axis is projected onto the surface.
When the optical power becomes zero, the outer part is projected.
If the height of the inner part on the surface is $h_p$, then that of its virtual image $h_v$ is computed as $h_p\frac{d_v}{d_p}$ according to geometric similarity, where $d_v$ is the distance from the ETL to the virtual image.
Because the outer part is observed when the ETL's optical power is zero, its real image is observed. 
Now, we suppose that an observer's eye is located at $d_e$ behind the ETL (Figs.~\ref{fig:lens-breathing-compensation}(a2) and \ref{fig:lens-breathing-compensation}(b2)).
Regarding the virtual image of the inner part, we denote the visual angle of its far end from the optical axis as $\theta_{i,f}$.
We also denote the visual angle of the near end of the outer part as $\theta_{o,n}$.
%Regarding the virtual image of the inner part, the visual angle of its far end from the optical axis is $\theta_{i,f}=\arctan\frac{h_v}{d_e+d_v}$.
%The visual angle $\theta_{o,n}$ of the near end of the outer part is $\theta_{o,n}=\arctan\frac{h_p}{d_e+d_p}$.
Figures~\ref{fig:lens-breathing-compensation}(a2) and \ref{fig:lens-breathing-compensation}(b2) depict the relationship between $\theta_{i,f}$ and $\theta_{o,n}$ for both cases where the non-zero optical power is positive and negative, respectively.
In general, when a larger optical power is applied to the inner part of a 3D CG object than the outer part, $\theta_{i,f}>\theta_{o,n}$, and, consequently, the retinal images of the inner and outer parts overlap.
On the contrary, when a smaller optical power is applied to the inner part, $\theta_{i,f}<\theta_{o,n}$, and a gap occurs between the two retinal images.

As discussed above, in a case where a 3D CG object is divided into multiple parts that are observed with different optical powers, an overlap or gap of their virtual images occurs on the observer's retina.
We compensate for these artifacts by resizing each part of the projected object.
The following method assumes that each part of the object is projected onto a planar surface that is perpendicular to the ETL's optical axis.
If a part of the object that is observed with the same optical power covers a non-planar surface, we further divide it into smaller parts until each surface area onto which the subdivided part is projected can be regarded as a plane that is perpendicular to the optical axis.
Suppose that $\theta_{p,n}$ is the visual angle of the near end of a part of a 3D CG object on the projection surface, and $\theta_{p,f}$ is that of its far end in a case where the object is not seen through an ETL (Fig.~\ref{fig:lens-breathing-compensation}(c1)).
Then, our method aims to resize the object so that the visual angles of the far and near ends of its virtual image correspond to $\theta_{p,n}$ and $\theta_{p,f}$, respectively.
Let the height of the near end and that of the far end be $h_{p,n}$ and $h_{p,f}$, respectively.
To achieve our goal, the heights of the near and far ends of its virtual image ($h_{v,n}$ and $h_{v,f}$, respectively) should be $h_{v,*}=\frac{d_v+d_e}{d_p+d_e}h_{p,*}$ according to geometric similarity, where $*$ represents either $n$ or $f$.
The ETL represents this virtual image as the projection of the displayed 3D CG object from the ETL's optical center onto a plane located at $d_v$ away from the ETL.
Therefore, to show this virtual image, the heights of the near and far ends of the 3D CG object on the surface should be $h'_{p,*}=\frac{d_p}{d_v}h_{v,*}$ according to geometric similarity (Fig.~\ref{fig:lens-breathing-compensation}(c2)).
Consequently, we can show the virtual image of the 3D CG object such that the visual angles of its far and near ends correspond to $\theta_{p,*}$ by resizing the object using the following equation:
\begin{equation}\label{eq:resize}
    h'_{p,*}=\frac{d_p(d_v+d_e)}{d_v(d_p+d_e)}h_{p,*}.
\end{equation}
An important consequence that is indicated by Eq.~\ref{eq:resize} is that the resizing factor is constant for the entire area of a subdivided part of the 3D CG object.

\subsection{Rendering pipeline}

We develop a new rendering pipeline for the projection images based on the conventional two-pass rendering technique~\cite{10.1145/280814.280861}.
The original algorithm consists of off-screen and on-screen rendering passes. % conducted in a virtual space where the positions of an observer and a projector as well as the shape of a projection surface are represented.
It requires the positions of an observer and a projector as well as the shapes of the projection surfaces and the 3D CG object to be rendered.
Next, (1) the off-screen rendering is performed to generate the desired appearance of the 3D object \revise{and its depth map} for the observer's viewpoint.
(2) The rendered image is then projected onto the surface from the observer's viewpoint in the virtual space.
(3) After that, the on-screen rendering is performed, in which the newly textured surface is captured from the projector's viewpoint.
(4) Finally, the captured image is projected in the physical space.
We incorporate the depth filtering and lens breathing compensation into the original two-pass rendering algorithm.

%修論図3.8
\begin{figure}[t]
  \centering
  \includegraphics[width=0.98\linewidth]{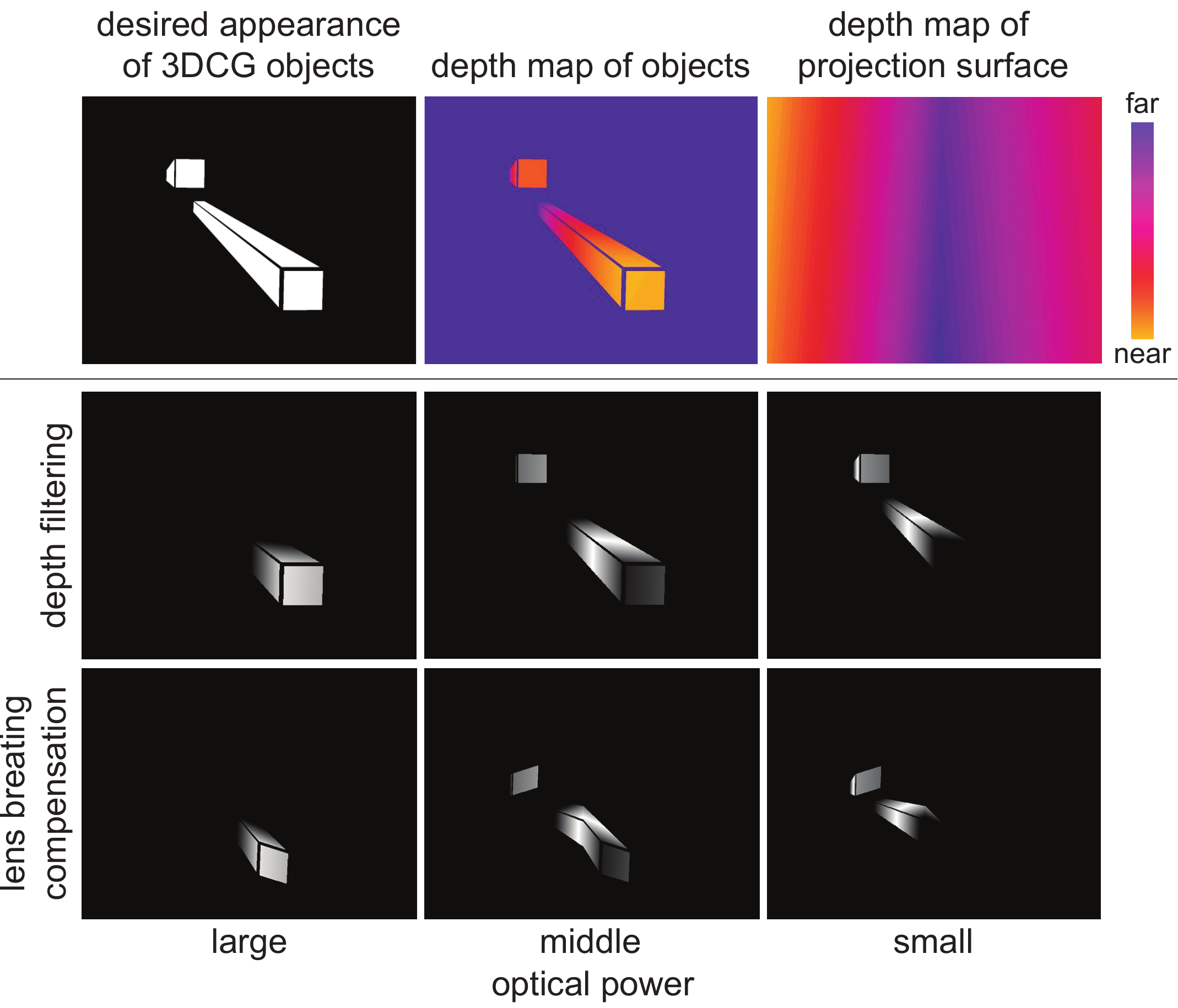}
  \caption{Generated images in the proposed rendering pipeline, where the target 3D CG objects were two cuboids of different heights and the physical projection surface was a corner-shaped screen.}
  \label{fig:rendering-pipeline}
\end{figure}

%修論図3.9
%\begin{figure}[t]
%  \centering
%  \includegraphics[width=0.98\linewidth]{figure/dummy.jpg}
%  \caption{FOV modification.}
%  \label{fig:fov-modification}
%\end{figure}

In our pipeline, the depth filtering is performed in the first process of the original algorithm.
%In addition, to render the 3D CG object, we obtain the depth maps of the object and the projection surfaces for the observer's viewpoint (the first row of Fig.~\ref{fig:rendering-pipeline}).
\revise{In addition to the rendering of the 3D CG object and its depth map, we obtain the depth map of the projection surfaces for the observer's viewpoint (the first row of Fig.~\ref{fig:rendering-pipeline}). Note that stereoscopic PM systems generally assume that the shape of a projection surface is obtained in advance and its pose is tracked online.}
We divide the rendered image into several parts using Eq.~\ref{eq:divide} such that each part is projected when the ETL's optical power becomes equal to a certain sampled value.
%We determine the ETL's optical power for each divided area of the rendered image using Eq.~\ref{eq:divide}.
As described in Sect.~\ref{subsec:method_vi}, the number of sampled optical powers is $N'$, and, thus, we render $N'$ images from the above process.
%The optical power for each divided part is determined using Eq.~\ref{eq:divide}.
We then apply depth filtering (Eq.~\ref{eq:depthf}) to these images (the second row of Fig.~\ref{fig:rendering-pipeline}).
Next, the lens breathing compensation is performed in the second process of the original two-pass rendering algorithm, in which we resize the images that are rendered in the first process.
We develop an efficient technique for the implementation of the resizing principle described in Sect.~\ref{subsec:method_comp}.
In particular, we found that the resizing can be performed just by modifying the FOV of the virtual projector in the second process because (1) the resizing is performed with respect to the optical axis of the eye, and (2) the same resizing factor is applied to each divided part of the 3D CG object (Eq.~\ref{eq:resize}).
Specifically, as shown in Fig.~\ref{fig:lens-breathing-compensation}(c2), the FOVs before and after modification are as follows:
\begin{equation}
    \theta_{p,f}=\arctan\frac{h_{p,f}}{d_e+d_p},\ 
    \theta'_{p,f}=\arctan\frac{h'_{p,f}}{d_e+d_p}.
\end{equation}
Therefore, we modify the FOV of the virtual projector by multiplying it by the factor of $\frac{\theta'_{p,f}}{\theta_{p,f}}$ when we project the rendered image in the second process of our two-pass rendering pipeline.
The third row of Fig.~\ref{fig:rendering-pipeline} shows an example of a rendered image after the lens breathing compensation has been applied.
%Figure~\ref{fig:fov-modification} shows the two FOVs before and after the modification in which the same variables are used as Fig.~\ref{fig:lens-breathing-compensation}.
%The visual angle of the far end of the rendered image is computed as follows:
%
%\begin{equation}
%    \theta_{p,f}=\arctan\frac{h_{p,f}}{d_e+d_p},
%\end{equation}
%
%where $d_e$ is the distance from the ETL to the eye.
%Suppose the distance from the optical axis to $y_{v,f}$ and that from the optical axis to $y'_{p,f}$ are $h_{v,f}$ and $h'_{p,f}$, respectively, then:
%
%\begin{equation}
%    h_{v,f}=(d_e+d_v)\tan\theta_{p,f},\ h'_{p,f}=\frac{h_{v,f}d_p}{d_v}.
%\end{equation}
%
%Finally, the visual angle $\theta'_{p,f}$ of $y'_{p,f}$, the far end of the resized rendered image is:
%
%\begin{equation}
%    \theta'_{p,f}=\arctan\frac{h'_{p,f}}{d_e+d_p}.
%\end{equation}
%
%Then, we modify the FOV of the virtual projector by multiplying the factor of $\theta'_{p,f}/\theta_{p,f}$ when we project the rendered image in the second process of our two-pass rendering pipeline.

\section{Experiment}

%We implemented a proof-of-concept prototype system and conducted several experiments to validate the proposed technique.
%First, we show the effectiveness of the lens breathing compensation technique in alleviating the overlap and gap artifacts.
%Second, we demonstrated multifocal PMs for different types of physical surfaces including a flat static surface, non-planar surfaces, and a moving surface.
%In particular, we validated the ability of our system to present physical objects as well as virtual objects in the experiment for the flat static surface.
%Finally, we conducted a user study of a depth matching task to investigate if VAC can be mitigated. 

\subsection{Experimental setup}\label{subsec:exp_setup}

%\begin{figure}[t]
%  \centering
%  \includegraphics[width=0.98\linewidth]{figure/dummy.jpg}
%  \caption{Experimental setup.}
%  \label{fig:experimental-setup}
%\end{figure}

We constructed a prototype system consisting of a pair of LC shutters, a pair of ETLs, and a synchronized high-speed projector (Fig.~\ref{fig:teaser}a).
We assembled our active-shutter glasses by attaching each ETL (Optotune AG, EL-16-40-TC, aperture: 16 mm, optical power range: -10 D to 10 D) onto an LC shutter taken from off-the-shelf active-shutter glasses (RV-3DGDLP1, field rate: 96-144 Hz) and inserted a pair of the combined device into an eyeglass frame fabricated using an FDM 3D printer to form a wearable (69$\times$128$\times$67 mm, 200 g) device.
The digital signal (sinusoidal wave) to control the ETLs was generated by a workstation (CPU: Intel Xeon E3-1225 v5 3.30GHz, RAM: 32GB, GPU: NVIDIA Quadro K620), input into a D/A converter (National Instruments, USB-6211), and converted into an analog voltage.
This voltage was then converted to an electric current through a custom amplifier circuit using an op-amp (LM675T).
Finally, the current was fed to the ETLs.
%According to the ETL's data sheet, the input analog voltages in our system ranged from -0.07 to 0.07 V.
The LC shutters were controlled using a circuit consisting of a motor driver (TOSHIBA, TA7291P).
We employed an off-the-shelf high-speed projector (Inrevium, TB-UK-DYNAFLASH, 1024$\times$768 pixels, 330 ANSI lumen) that can project 8-bit grayscale images at 2,000 frames per second (fps)\footnote{Although the frame rate on the specification sheet is 1,000 fps, we experimentally discovered that the projector could project grayscale images at 2,000 fps.}.
The projection images were generated by the workstation and sent to the projector via a PCI Express interface.
\revise{We manually adjusted the projector location so that the projected result always appeared focused on a projection surface. We manually measured the surface's shape and its pose relative to the ETLs offline. In Sect.~\ref{subsec:exp_mov}, a surface was tracked using a motion capture system consisting of five cameras (NaturalPoint, OptiTrack Prime 17W). The shape and pose information was used to obtain the depth map in our rendering pipeline.}

%修論図4.1
\begin{figure}[t]
  \centering
  \includegraphics[width=0.93\linewidth]{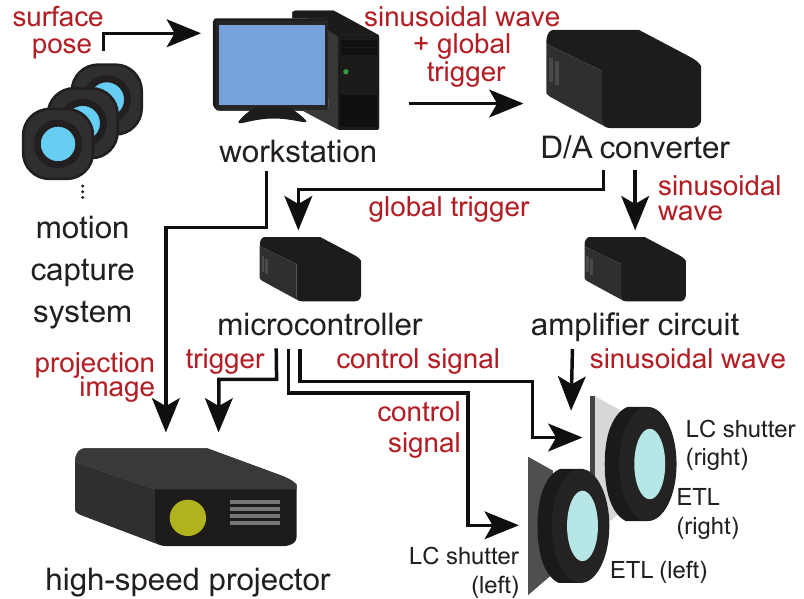}
  \caption{System architecture.}
  \label{fig:system-architecture}
\end{figure}

The workstation was able to generate a trigger signal for the projector and a control signal for the LC shutters along with the sinusoidal wave for the ETLs.
However, the D/A converter only had two output channels, which was not sufficient for sending all the signals.
Therefore, we used an additional microcontroller (Arduino Uno) to synchronize the display timing of each projection image and the opening/closing timing of the LC shutters.
In our system, the workstation sent the sinusoidal wave $i[\phi]$ and a single global trigger signal at $\phi=0$ to the D/A converter.
The trigger signal was then transferred to the microcontroller, which generated the trigger and the control signal using the delays of $\delta t_p$, $\delta t_{c2o}$, and $\delta t_{o2c}$, and sent to the projector and the controlling circuit of the LC shutters.
The system architecture is depicted in Fig.~\ref{fig:system-architecture}.
%We assume that our system works in a dark environment.

\subsubsection{System calibration}

%修論図4.3, 4.4
\begin{figure}[t]
  \centering
  \includegraphics[width=0.98\hsize]{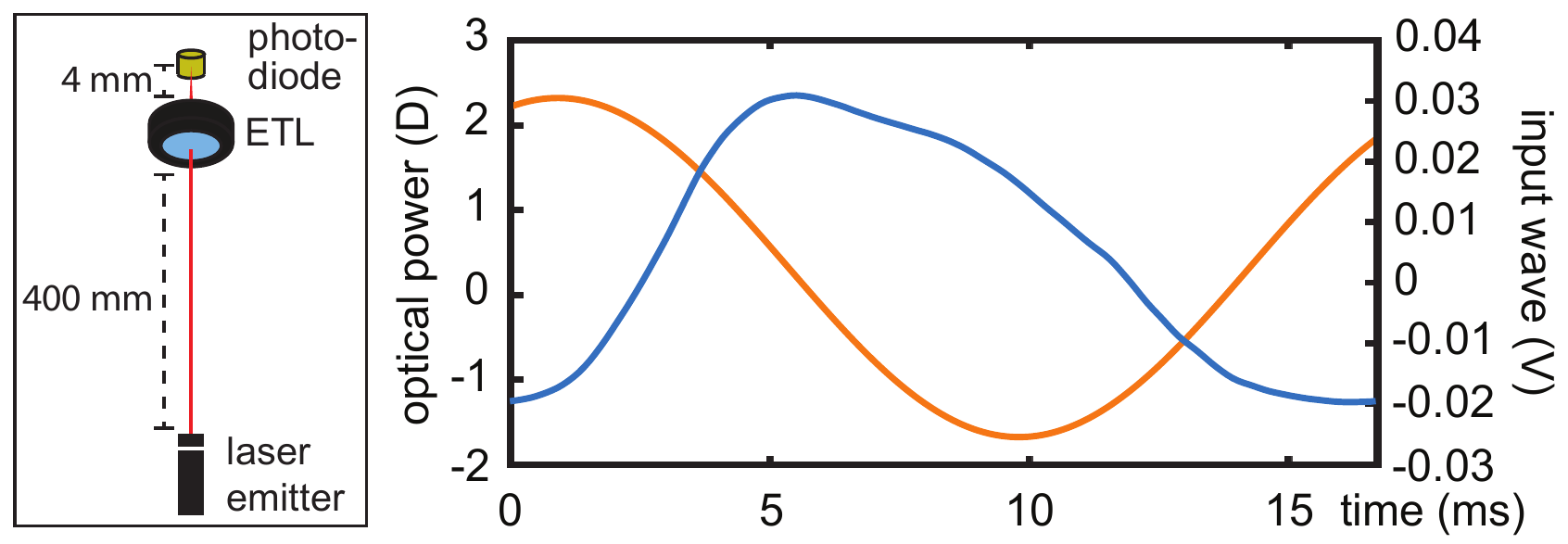}
  \caption{Optical power measurement: (left) experimental setup, (right) one cycle of input sinusoidal wave (orange line), and the resultant optical power of an ETL (blue line).}
  \label{fig:etl-calibration}
\end{figure}

The system calibration was conducted as follows.
We built an optical power measurement system using a photodiode and a laser emitter (wavelength: 635 nm, output power: 0.9 mW) (Fig.~\ref{fig:etl-calibration}).
We placed the ETL at about 400 mm away from the laser emitter.
The photodiode was placed about 4 mm behind the ETL, and it measured the laser beam passing through the ETL.
Raising the optical power of the ETL reduced the laser's spot size on the photodiode (i.e., increased the laser power density), enhancing the electric current created in the photodiode.
We indirectly measured the ETL's optical power in this manner.
For the calibration of the ETL, we input a sinusoidal wave $i[\phi]$ to the ETL and measured one cycle of its output optical power $v[\phi]$.
The frequency of the wave was maintained at 60 Hz throughout the experiment, which was higher than the CFF.
According to the discussion in Sect.~\ref{subsec:method_vi}, the sweep range was determined based on the desired location range of the virtual images (i.e., $d_{v,n}$ and $d_{v,f}$) on the nearest physical surface from the ETL.
%In the following experiments, except for the user study in Sect.~\ref{subsec:exp_user}, the nearest physical surface was placed 500 mm away from the ETL, and $d_{v,n}=$ 167 mm and $d_{v,f}=$ $\infty$ mm.
%Therefore, the required sweep range was between -1.0 D and 2.0 D.
\revise{In the following experiments, except for the user study in Sect.~\ref{subsec:exp_user}, we considered the nearest distance as 500 mm, and $d_{v,n}=$ 167 mm and $d_{v,f}=$ $\infty$ mm. Therefore, the required sweep range was between -1.0 D and 2.0 D according to Eq.~\ref{eq:thin_lens}. We can still modulate the location of the virtual image of a projected 3D CG object onto a surface whose distance from the ETL is shorter than 500 mm. But in such a case, the modulation range becomes narrower than that between $d_{v,n}$ and $d_{v,f}$.}
We manually searched for the offset and amplitude of the input sinusoidal wave by which the output optical power of the ETL can cover the target range and found the suitable values (offset: 2.5 mV, amplitude: 28 mV).
Figure~\ref{fig:etl-calibration} shows the measured optical power $v[\phi]$ by the input wave.
From the figure, we found that the output values did not form a clean sinusoidal wave.
Therefore, we recorded one cycle of the output wave along with the corresponding input wave and stored it on the microcontroller to be able to look up the optical power at a given phase of the input wave.

%本システムでは，予めETLの入力波形に対する屈折力変化を記録し，プロジェクタ投影タイミングと液晶シャッタ開閉タイミングを決定した．液晶シャッタもレーザ光線とフォトダイオードを用いて開閉の立ち上がり時間と立下り時間を計測した．立ち上がり時間は約3.0ms で，立下り時間は約0.1 ms だった．また，プロジェクタのトリガ入力から投影までの遅延は約164 μ s だった．DACは正弦波と同時にトリガを同じ60 Hz でマイコンボードに出力する．DAC からのトリガと屈折力を予め計測し，DAC からのトリガを基準としてETL が所望の屈折力になる時間を計算する．マイコンボードは，そのデータを基にDAC からのトリガを基準にして所望の屈折力でプロジェクタ投影と液晶シャッタ開閉が行われるよう，それぞれにトリガ信号を送る．

To synchronize the ETLs and the high-speed projector, we used the same photodiode to measure the delay for the high-speed projector between a trigger signal being sent from the microcontroller and the actual projection.
As a result, we found that the delay was 0.15 ms ($=\delta t_p$). % Ueda's data: 0.46 ms
We also measured the delay for the LC shutters and found that the closing transition took 0.1 ms ($=\delta t_{o2c}$), while the opening transition took 3.0 ms ($=\delta t_{c2o}$).
This delay information was also stored in the microcontroller.
The microcontroller received the global trigger signal, computed when the trigger and control signals should be sent to the projector and LC shutters based on the recorded optical power waveform and the delay information, and, finally, sent the signals to the devices.
%Using this delay information and the data in the database, we can use to projector to illuminate a real object exactly when the ETLs' optical powers are the target optical power.

%Projector2(Texas Instruments, DLP LightCrafter 4500, 912x1140)
%using OpenGL/GLSL to generate image

\subsection{Lens breathing compensation}\label{subsec:exp_comp}

We verified whether the proposed lens breathing compensation improved the gap or overlap between adjacent areas of a projected 3D CG object, which were observed with different optical powers.
The ETL was attached to a camera (Sony $\alpha$7S) with a suitable lens (Sony SEL24F14GM, focal length: 24 mm) that was used to capture the experimental results with a shutter speed of 1/60 seconds.
We placed a flat screen at 500 mm away from the ETL such that it was perpendicular to the ETL's optical axis.
The displayed 3D CG object was a textured plane.
The virtual image of the object was placed such that the distance of the left edge from the ETL was 333 mm, and that from the right edge was 1666 mm.
The attached texture was a black and white checker pattern.
We projected each part of the 3D CG object when the ETL's optical power was either -1.0 D, -0.5 D, 0.0 D, 0.5 D, 1.0 D, or 1.5 D.
%The depth filtering was applied to display some areas of the virtual object which should be projected in between the above optical powers.
We captured the projected result under two conditions: when the lens breathing compensation was (1) applied and (2) not applied.

%修論図5.4。ただし、depth filteringありなしは比較しない。
\begin{figure}[t]
  \centering
  \includegraphics[width=0.98\hsize]{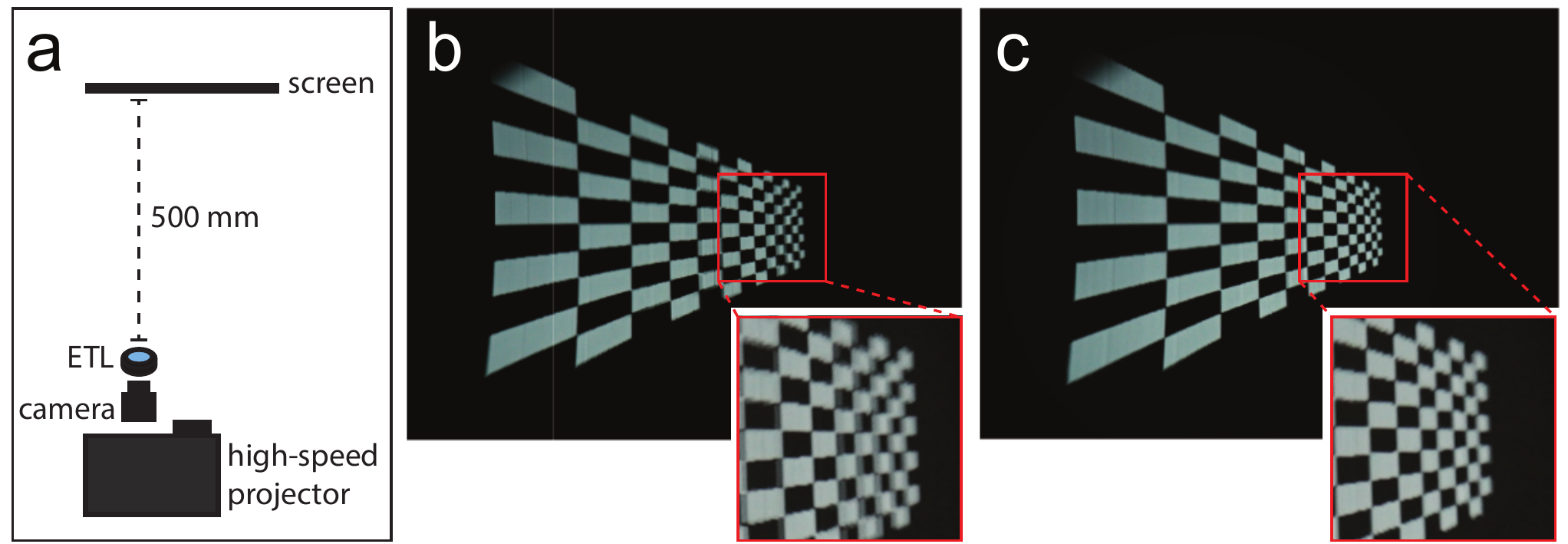}
  \caption{Experimental validation of the proposed lens breathing compensation technique. The projected 3D CG object was a slanted plane with a checker pattern. (a) Experimental setup. The object's appearances without (b) and with (c) the compensation.}
  \label{fig:exp-lens-breathing}
\end{figure}

Figure~\ref{fig:exp-lens-breathing} shows the results.
Comparing the results of the two conditions, we found that the appearance of the checkerboard texture significantly improved when the proposed compensation method was applied.
In particular, the unnatural overlaps of the textures degraded the image quality of the projected result without the compensation, while such artifacts were almost not visible in the result with the compensation.
Therefore, we confirmed that the proposed lens breathing compensation technique works well to alleviate the artifacts caused by the lens breathing phenomenon.

\subsection{Multifocal appearance of a projected object on a flat static surface}\label{subsec:exp_static}

%修論図5.1と図5.2
%図5.2の(ETL Fixed-No sync, Focused at 2D)をベースラインとして見せる ?
%同図の(ETL Sweep-Sync, 全て)を提案法の結果としてみせる。つまり、全部で4枚載せる
%\begin{figure}[t]
%  \centering
%  \includegraphics[width=0.98\hsize]{figure/figure-multifocus-on-static-su%rface-2.png}
%  \caption{Multifocal projection onto a flat static surface. (note: let %change the layout of figure later)}
%  \label{fig:exp-static}
%\end{figure}

We validated our technique using a flat, static screen.
As shown in Fig.~\ref{fig:teaser}b, we placed the screen 500 mm away from the ETL, which was attached to the same camera mentioned in Sect.~\ref{subsec:exp_comp}.
In addition, three physical objects were placed at different distances from the ETL, which were 333 mm (near), 500 mm (middle), and 1,000 mm (far).
We projected three Stanford bunnies on the screen such that the distance of the virtual image of each bunny from the ETL was the same as one of the physical objects.
Specifically, the near, middle, and far bunnies were projected when the ETL's optical power was -1.0 D, 0.0 D, and 1.0 D, respectively, which were computed using Eq.~\ref{eq:thin_lens}.
The near, middle, and far bunnies were displayed from the bottom to top of the screen.
The physical objects were illuminated by a uniform white image when the optical power was 0.0 D.
We captured the scene using the camera by changing its focus distance from 333 mm to 1,000 mm.
%As a base line, we also captured a conventional projection mapping scene using the camera with the focusing distance of 500 mm.
%In this scene, the projector projected all the three bunnies onto the screen and a uniform white image onto the physical objects while the focal sweep was stopped and the ETL's optical power was fixed as 0 D.

Figures~\ref{fig:teaser}c, \ref{fig:teaser}d, and \ref{fig:teaser}e show the results.
When the focus distance of the camera was 333 mm, the physical object that was placed in the near position appeared more focused than the other physical objects.
Similarly, the image of the bottom bunny, which was projected when the optical power was -1.0 D, appeared more focused than that of the other projected bunnies.
We obtained the same observations when the camera's focus distance was 500 mm and 1,000 mm as well.
Therefore, we concluded that the proposed method could provide a multifocal view in PM while presenting the physical objects without any focus modulations.

\subsection{Multifocal appearance of a projected object on a non-planar surface}\label{subsec:exp_nonp}

%段差スクリーン実験
\begin{figure}[t]
  \centering
  \includegraphics[width=0.98\hsize]{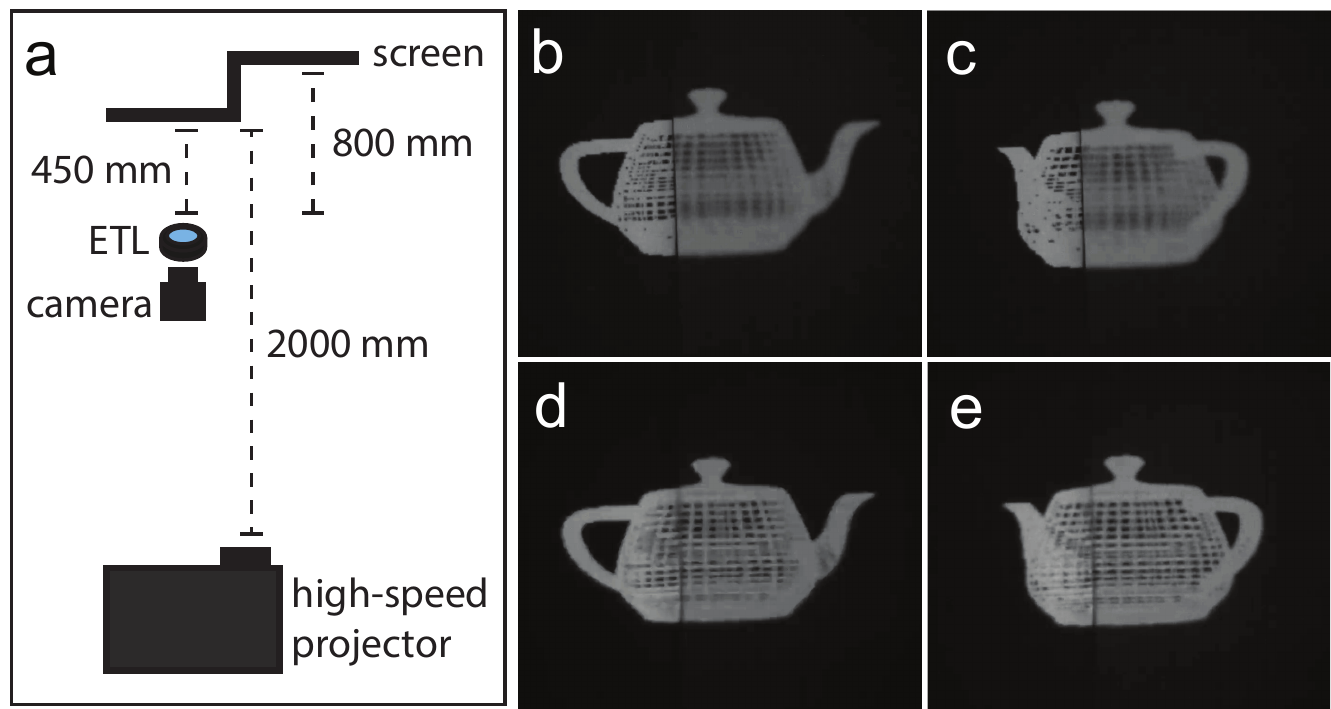}
  \caption{Multifocal appearance of a rotating Utah teapot projected onto a step surface. (a) Experimental setup. (b, c) The projected object observed through the ETL with a fixed optical power of zero. (d, e) The projected object observed using the proposed method.}
  \label{fig:exp-non-planar}
\end{figure}

A unique technical challenge when attempting to mitigate the VAC in stereoscopic PM is that the display surface can potentially be non-planar.
Thus, we verified whether the proposed system could provide the correct focus cues when the projection screen is non-planar.
We placed a step surface such that the two planes were facing the ETL.
The distance of the planes from the ETL were 450 mm and 800 mm, respectively.
%Figure~\ref{fig:exp-non-planar} shows the physical surface used as the screen in this experiment.
%The screen consisted of the top of a box and a wall behind it, and located between xxx and xxx mm from the ETL attached on the same camera used in Sect.~\ref{subsec:exp_comp}.
We projected a Utah teapot such that it covered the two planes.
\revise{We placed the projector about 2.0 m away from the surface, which caused the projected images to be focused on the both planes.}
Using our technique, the virtual image of the object was positioned such that the center of the object was at the same distance from the ETL as the front plane.
We projected each part of the object when the ETL's optical power was either -1.0 D, -0.5 D, 0.0 D, or 0.5 D.
As a baseline, we also captured the object while keeping the optical power fixed at 0.0 D.
Therefore, there were two conditions for which the scene was recorded: (1) with and (2) without our technique.

Figure~\ref{fig:exp-non-planar} shows the results captured by a camera (Ximea MQ013CG-ON) with a C-mount lens (focal length: 12 mm) to which the ETL was attached.
Comparing the results of the two conditions, we found that all parts of the teapot observed using our technique appeared focused when we adjusted the camera lens such that the focus point corresponds to the front projection plane.
On the other hand, the right side of the teapot appeared blurred in the condition where our technique was not used.
%, in which we used the same focusing distance.
Thus, we concluded that the proposed technique could successfully realize multifocal PM for a non-planar screen.

\subsection{Multifocal appearance of a projected object on a moving surface}\label{subsec:exp_mov}

%修論図5.7 と 5.8
\begin{figure}[t]
  \centering
  \includegraphics[width=0.98\hsize]{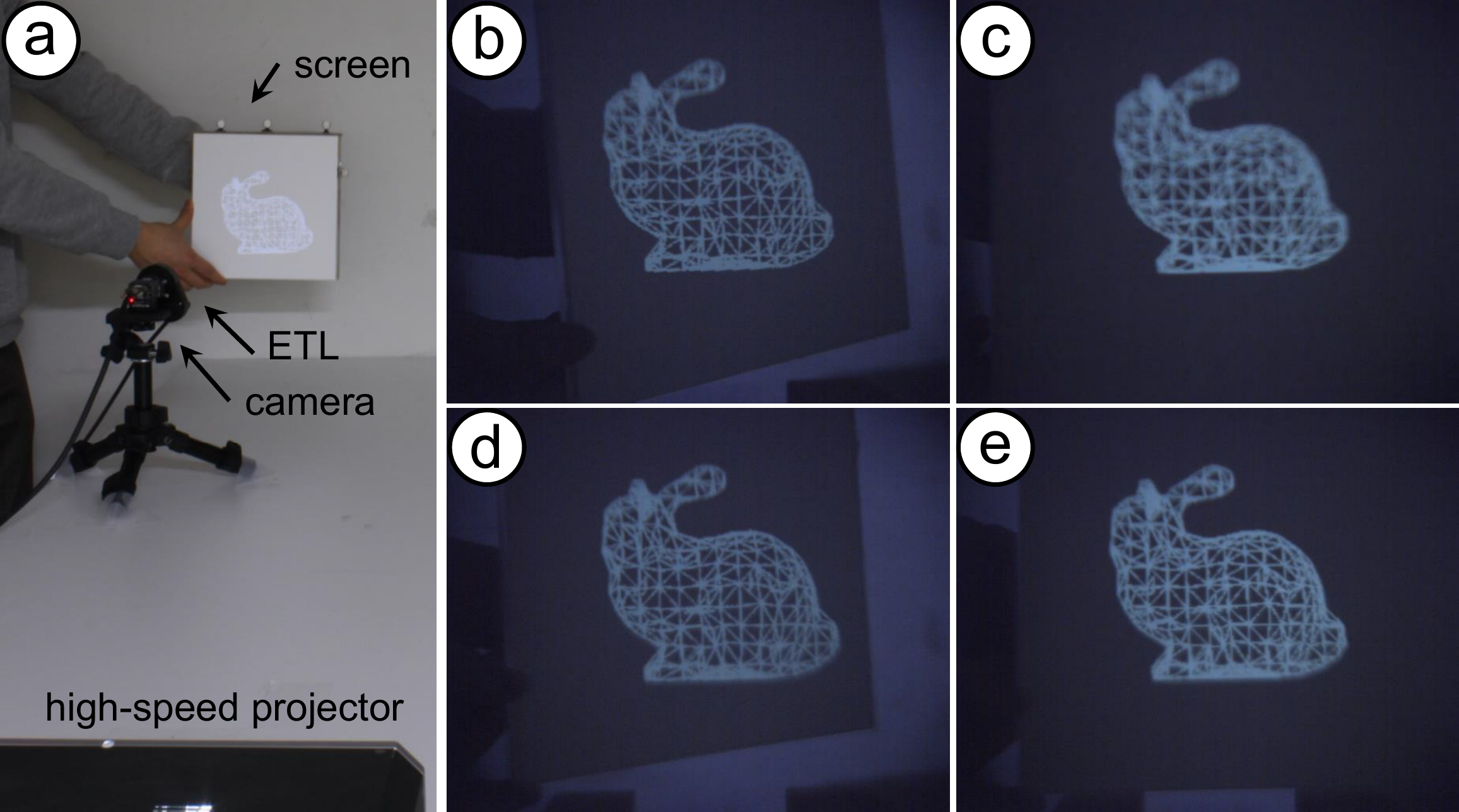}
  \caption{Multifocal appearance of a Stanford bunny projected onto a flat surface freely moved by hand. (a) Experimental setup. (b, c) The projected object observed through the ETL with a fixed optical power of zero. (d, e) The projected object observed using the proposed method. \revise{Need revision}
}
  \label{fig:exp-moving}
\end{figure}

Another unique technical challenge when attempting to mitigate the VAC in stereoscopic PM is that the display surface can potentially move.
Thus, we verified whether the proposed system could provide the correct focus cues when the projection screen moves in an experimental setup, shown in Fig.~\ref{fig:exp-moving}.
A flat screen was manually moved between 300 mm and 500 mm from the same camera mentioned in Sect.~\ref{subsec:exp_nonp}.
%ETL attached on a camera (Ximea MQ013MG-ON) with a C-mount lens (focal length: 12 mm) by which we recorded the virtual image of a projected 3D CG object.
The projected object was a Stanford bunny such that the 3D position of its virtual image in the scene was fixed regardless of the screen pose.
We controlled the timing of the projection so that the bunny always appeared focused.
The bunny image was projected when the ETL's optical power was either -0.5 D, 0.0 D, 0.5 D, 1.0 D, or 1.5 D.
As a baseline, we also recorded a video of the bunny while keeping the optical power fixed at 0.0 D.
Therefore, the same two experimental conditions as discussed in Sect.~\ref{subsec:exp_nonp} were established.
As indicated in Sect.~\ref{subsec:exp_setup}, the screen pose was tracked using a motion capture system.
%To track the screen pose, we used an off-the-shelf motion capture system consisting of five cameras (NaturalPoint, OptiTrack Prime 17W).

Figure~\ref{fig:exp-moving} shows the results that were captured at different screen poses.
First, we ensured that the size and position of the bunny were consistent throughout the video-recording process in both experimental conditions.
%of with and without our technique.
This indicates that our two-pass rendering pipeline worked correctly.
Comparing the results of the two conditions, we found that the bunny observed using our technique always appeared focused, while the one projected without our technique appeared blurred at certain screen poses.
Thus, we concluded that the proposed technique could successfully realize multifocal PM for a moving screen.

\subsection{User study of depth-matching}\label{subsec:exp_user}

We conducted a user study to investigate the effect of the VAC on the depth estimation accuracy of a human observer in stereoscopic PM.
Previous studies have revealed that the VAC caused distortions in the perceived depth; more specifically, people overestimated the depth of an AR target when using conventional OST displays~\cite{7164348,8462792,9284699}.
Our user study investigated the depth estimation accuracy in a depth-matching experiment that was designed based on the previous studies.
Each participant moved a physical pointer to match its depth with a virtual target displayed using stereoscopic PM.
We prepared two experimental conditions.
The first one was the ``proposed'' condition, where the VAC was mitigated by our system, and the second one was the ``conventional'' condition, where the VAC was not mitigated and only the binocular cue was correct.
Based on the results of the previous studies~\cite{7164348,8462792,9284699}, the hypotheses of our study were formulated as follows:
\begin{description}
    \item[H1:] The depth was overestimated when the VAC was not mitigated in stereoscopic PM.
    \item[H2:] The depth estimation was more accurate when the VAC was mitigated by our proposed system.
\end{description}
%

%修論図6.5, 6.2, 6.7, 6.8
\begin{figure}[t]
  \centering 
  \includegraphics[width=0.98\hsize]{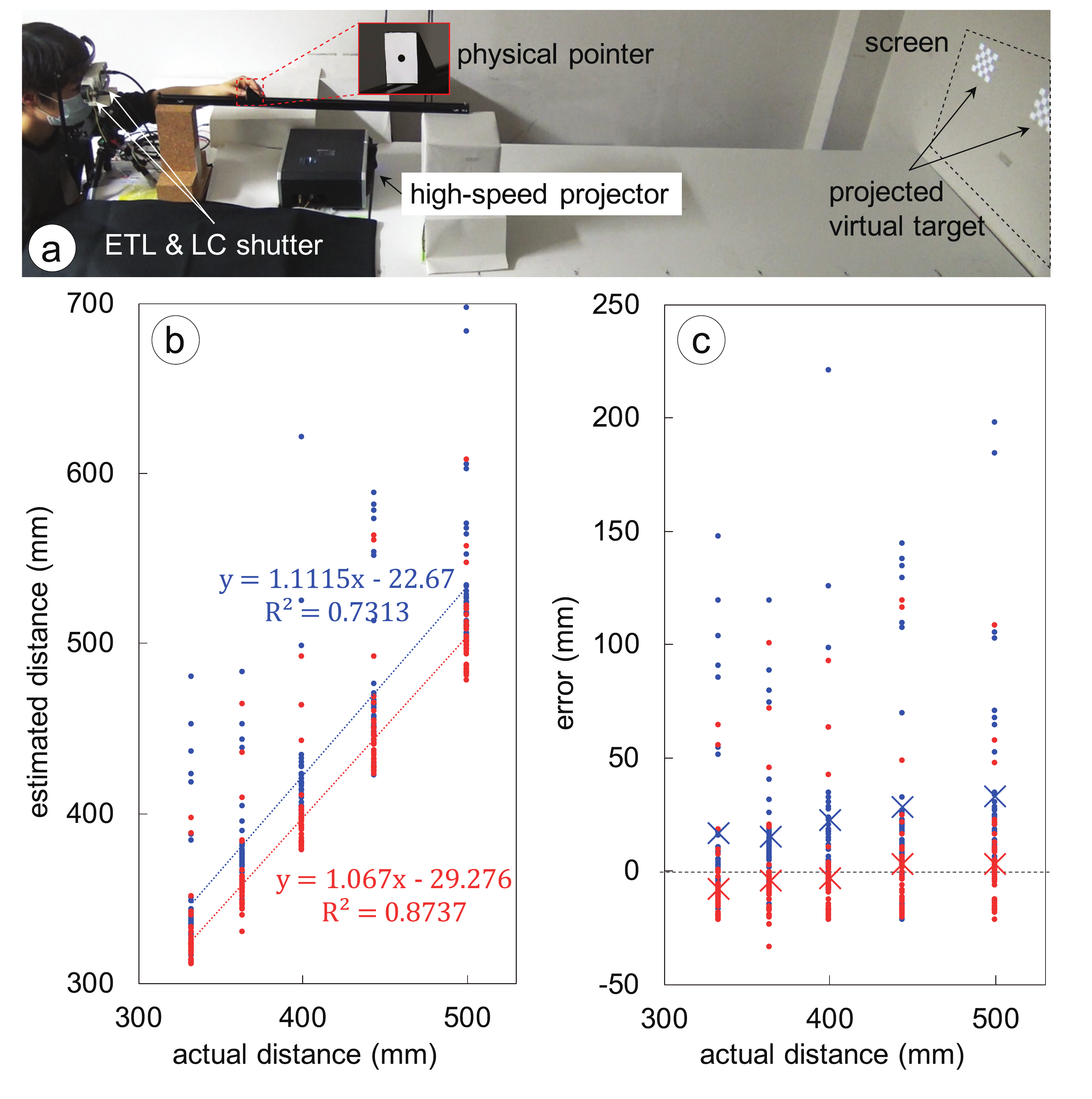}
  \caption{Depth-matching experiment. (a) Experimental setup where a participant manually moved a physical pointer to match its depth with a projected virtual target (checker pattern). Note that the study was conducted under a dark environment. (b) The scatter plot of the estimated against the actual distance with fitted linear equations. The red and blue plots represent the proposed and conventional conditions, respectively. (c) Errors of the estimated from the actual distance in both conditions; ``x'' indicates the mean value.}
  \label{fig:exp-user-study}
\end{figure}

\subsubsection{Experimental setup and procedure}

We built an experimental system in a dark room, as shown in Fig.~\ref{fig:exp-user-study}a.
Participants sat on a chair and placed their faces on a chin rest.
We attached our active-shutter glasses consisting of the ETLs and the LC shutters to the chin rest such that the participants could see the experimental scene through our glasses.
The virtual target was a checker pattern, which was projected onto a flat screen placed 2.5 m away from the ETLs.
The visual angle of the target was fixed at 2.4 degrees throughout the experiment.
By providing the appropriate disparities to both eyes, our system presented the virtual target at five distances from the ETLs: 300 mm, 364 mm, 400 mm, 444 mm, and 500 mm.
The sweep range of the ETL's optical power in this experiment was -2.6 D to 0.0 D.
Under the proposed condition, the virtual target was projected when the ETL's optical power was -2.6 D, -2.1 D, and -1.6 D for the target distances of 300 mm, 400 mm, and 500 mm, respectively.
We applied the depth filtering technique to present the focus cues for the other two target distances.
For the conventional condition, the virtual target was projected when the optical power was zero for all the target distances.
For both conditions, the physical pointer was attached to a linear slider using which a participant could move the pointer along the direction of the ETL's optical axis.
The system illuminated the physical pointer when the ETL's optical power was zero using a synchronized light.
%The physical pointer was moved below the ETL's optical axis while the virtual target was projected above it to avoid the interference between them.

We recruited ten participants from a local university.
All participants were na\"{i}ve to the purpose of the experiment and had normal or corrected-to-normal vision.
We divided the participants into two groups; half of them performed depth-matching tasks under the proposed condition on the first day and did the same under the conventional condition on the second day, while the other half performed the tasks in the opposite order.
In a preparation block, we measured the interpupillary distance (IPD) for each participant and accordingly adjusted the distance between the two ETLs.
Then, we asked them to adjust their head positions by observing a calibration pattern to ensure that the disparities of the physical pointer and the virtual target were identical at the same distance.
Specifically, we projected a cross pattern onto the screen for each eye and positioned a physical object on which the same cross pattern was printed such that all the crosses corresponded to the participant's retina when their head position was correct.
After the preparation block, each participant performed the following depth-matching tasks, including a training task.
In each task, they were asked to move the physical pointer by hand until they perceived that it was located at the same depth as the virtual target.
We prepared five depths as described above, and each participant performed the depth-matching tasks four times for each depth.
Thus, there were 20 ($=5$ depth $\times$ 4 repetitions) trials in addition to the training trial for each condition.
The order of the trials was counterbalanced among the participants.
Further, the participants were allowed to take as much time as they needed to complete each task.
We told them that they could give up on a task if they could not perform the binocular fusion.
After all the tasks were completed, we asked each participant whether they perceived the flicker of the displayed object.

\subsubsection{Result}

All the participants were able to perform the binocular fusion in all the tasks and reported that they did not perceive any flickers during the experiment.
%In addition, there was no trial where the participants could not perform the binocular fusion.
%Therefore, we concluded that our system appropriately provided the binocular cue in the both conditions.
Figure~\ref{fig:exp-user-study}b shows the results plotted as a scatter plot of the estimated against the actual distance.
Following previous studies~\cite{7164348,8462792,9284699}, we calculated the linear functions that predict the estimated distance from the actual distance and used multiple regression methods to determine whether the slopes and intercepts significantly differed between the conditions.
%It was shown in~\cite{7164348} that the multiple regression method is more suitable than ANOVA, because the slopes and intercepts yielded by the multiple regression are descriptive statistics and more useful than means, because they directly describe functions that predict estimated distances from actual target distances.
While Fig.~\ref{fig:exp-user-study}b shows all 400 data points ($=20$ trials $\times$ 2 conditions $\times$ 10 participants), we averaged the responses for the four repetitions for each participant, which reduced the size of the dataset to 100 points for the multiple regression analysis~\cite{7164348}.
We then compared the two linear fits of the proposed and conventional conditions.
Our multiple regression analysis found that the slopes of the linear fits did not significantly differ ($F_{1,96} = 0.218, p = 0.64$), while 
%Our analysis then tested whether the intercepts of the linear fits significantly differed, in which the slopes of the linear fits were adjusted to a common value.
the intercepts did ($F_{1,97} = 19.552, p < 0.01$).
%Therefore, the estimated distances in the proposed condition were best fit by the equation of $y = 1.08926x-38.3678$ mm and those in the conventional condition were best fit by the equation of $y = 1.08926x-13.5778$.
%Therefore, the depth estimation with VAC resulted in a constant 24.8 mm of overestimation relative to that with VAC mitigation.

Figure~\ref{fig:exp-user-study}c illustrates the errors of the estimated distance from the actual distance, where positive errors indicate the overestimation.
The participants overestimated the depth by 22.86 mm on average in the conventional condition, while the error was much smaller (-1.93 mm) in the proposed condition.
We performed a two-way analysis of variance (ANOVA) for the actual distance of the virtual target and the experimental condition to evaluate the estimation errors.
Again, we averaged the responses of the four repetitions for each participant.
The ANOVA showed the main effect of the experimental condition ($F_{1,90}=18.234, p<0.01$), but it did not show the main effect of the actual distance of the virtual target ($F_{4,90}=0.861, p=0.49$) and the interaction effect ($F_{4,90}=0.088, p=0.99$).
Next, a post-hoc analysis was performed using Tukey's HSD method for pairwise comparison.
The result showed a statistically significant difference between the proposed and conventional conditions ($p<0.01$).

Therefore, both our hypotheses were supported by the results.
In stereoscopic PM, the participants overestimated the depth of the virtual target when the VAC was not mitigated (H1) and estimated it more accurately when it was mitigated (H2).
These trends are consistent with the previous results of OST-AR displays~\cite{7164348,8462792,9284699}.

\section{Discussion}

From the experimental results, we confirmed that our multifocal stereoscopic PM technique successfully worked for different types of projection screens.
Specifically, it could provide the correct focus cues for static (Sect.~\ref{subsec:exp_static}), non-planar (Sect.~\ref{subsec:exp_nonp}), and  moving surfaces (Sect.~\ref{subsec:exp_mov}).
We also confirmed that our lens breathing compensation technique can be used to significantly reduce the associated artifacts (Sect.~\ref{subsec:exp_comp}).
Further, because it is a general solution, it can be applied to other AR/VR multifocal and varifocal displays and improve their image qualities.
Furthermore, the user study showed that the proposed technique could mitigate the VAC and significantly reduced the distortions in perceived depth in stereoscopic PM (Sect.~\ref{subsec:exp_user}).

%Spatial ARの枠組みで考察してみる。比較はholo（Oliverのやつ）, light field (gordon, debevecの昔の回転してるやつと、naganoさんとのやつ）, stereoscopic PMの三つ。holography、light fieldはいずれも原理上は正しいfocus cueもbinocular cueも出せるところが利点。巨大化しようとすると、light fieldも含め、大きなoptical element, phase modulatorが必要になる。一方、stereoscopic PMは大きな映像提示が可能で、予め環境側に設置すべきものがない。
%Stereoscopic PM is an implementation variation of spatial AR (SAR) which exploits spatially-aligned optical elements and displays~\cite{10.5555/1088894}.
%Light field~\cite{10.1145/2782782.2792494,10.1145/2601097.2601144} and holographic~\cite{1319281} displays are other major implementations of SAR.
%Both theoretically provide correct focus cues as well as binocular cues.
%On the other hand, these approaches require optical elements such as an anisotropic diffuser and hologram to be installed in the environment, which potentially interfere with user's activity especially when a large 3D CG object is displayed.

\subsection{Limitations}

Although we achieved the goal of this research, our technique still suffers from several technical limitations.
First, the current system works properly only in a dark environment because \revise{a bright environment increases the black offset of a projected result and, consequently, degrades its contrast}.
% because it requires rigorous illumination control.
Other PM systems also suffer from the same limitation because the environment lighting decreases the contrast of the projected imagery~\cite{10.5555/1088894}.
Previous attempts have addressed the limitation by illuminating the entire scene surface using multiple projectors, which can theoretically reproduce the original environment lighting~\cite{10.1145/2642918.2647383}.
%Our system also works in such ``ubiquitous projection'' environments.
% tsukamoto \cite{7164338}
In addition, by using large-aperture optics~\cite{8798245}, the environment lighting can be naturally reproduced.
\revise{A potential extension of our system is to apply anisotropic diffusers or retro-reflectors as projection screens~\cite{Eldes:13,10.1145/1836821.1836841}, but this, on the other hand, limits the scalability of the system.}

Second, the optical power of the ETL changes during each projection frame displayed because it is continuously modulated over time.
%Second, even when the system projects 3D CG objects that should be presented with the same optical power, the optical power of the ETL will change.
In addition, a certain amount of time is required for the projection of a 3D CG object to provide sufficient light intensity for an observer.
%Therefore, this problem occurs even in the case where our system is perfectly synchronized, and each part of the object is projected exactly when the ETL presents a desired optical power.
As a result, an observer perceives the integration of the undesirable appearances of a projected image.
Other multifocal displays that employ the focal sweep approach (see Sect.~\ref{subsec:rw_vac}) suffer from the same limitation.
Using a custom high-speed projector, we can illuminate the scene for a shorter duration to mitigate the image quality degradation, but this results in the image having a darker appearance.
Another and more promising solution for overcoming this limitation is to apply a waveform optimization of ETL's drive signal~\cite{Iwai2019}, which will improve the image quality by maintaining the optical power at the desired values for a sufficient amount of time.

Third, as an implementation limitation, our system requires a high-speed projector.
Although such a projector can currently be considered special equipment, the one we used in this paper is an off-the-shelf projector that is available in the market~\cite{watanabe2015high}.
In addition, researchers have found in the last few years that a high-speed projector is essential for realizing immersive dynamic PM applications~\cite{10.1145/3415255.3422888,8998378,9284678}.
Although our current system applies a gray-scale projector, the manufacturer of the projector just launched a full-color version.
Given this technical trend, high-speed projectors will probably soon become normal equipment in PM applications.

Foruth, because a projector generally suffers from a narrow depth of field (DOF), a part of the image projected on a non-planar surface does not appear focused in some cases.
Here, even if the proposed technique successfully displays the virtual image of a projected object at the desired distance, the virtual image does not appear focused.
%even when a user adjusts the optical power of their eyes to focus on it.
This can be solved by attaching another ETL to the projector and applying a focal sweep to it \cite{10.1117/12.2542477,7014259}.

\revise{Last, the ETL significantly reduces the angle of viewing for an observer. This problem can be also seen in previous AR/VR displays that employ ETLs~\cite{10.1145/2858036.2858140,8458263}. Considering the trend of the increasing aperture size of ETLs due to the constant innovation of manufacturers, we believe that this limitation will be overcome in the near future.}

In summary, although there are currently several technical limitations that are common to the related fields, we believe that they do not immediately degrade the applicability of the proposed method.

\section{Conclusion}

This paper aimed at achieving multifocal stereoscopic PM to improve the mismatch between vergence and accommodation.
To achieve this objective, we jointly designed the optics, hardware, and computational algorithms.
Specifically, we attached ETLs to the LC shutters of active-shutter glasses to control both vergence and accommodation and synchronized them using an off-the-shelf high-speed projector.
We applied a fast focal sweep to the ETLs and projected a 3D CG object from the projector onto physical surfaces when the distance of its virtual image was located exactly at a desired distance from the ETLs.
Further, we solved three technical issues that are unique to the stereoscopic PM and developed a simple lens breathing compensation technique that can be applied to other multifocal and varifocal AR/VR displays.
%a unique issue in the stereoscopic PM---the 3D CG object is displayed on non-planar and even moving surfaces---explicitly taking the distance of a physical surface from the ETL into account when we determine the ETL's optical power.
Using a proof-of-concept prototype, we have demonstrated that our technique can provide the correct focus cues required for different types of projection screens.
We have also confirmed that the lens breathing compensation can eliminate artifacts such as overlaps between the adjacent areas of projected imagery.
Finally, we validated the advantage provided by our technique with regard to the mitigation of the VAC by comparing it with conventional stereoscopic PM using a user study of a depth-matching task.
%Our future works are solving the above-discussed technical limitations.
\revise{A new branch of research associated with AR/VR displays has recently been enabled by the combination of projectors and near-eye optics~\cite{9383112,8999805,Ueda:21}, and this work can be categorized with the same. We will continue to investigate the possibility and applicability of this emerging framework.}

\acknowledgments{
This work was supported by JSPS KAKENHI Grant Numbers JP18K19817 and JP20H05958 and JST, PRESTO Grant Number JPMJPR19J2, Japan.
}

\bibliographystyle{abbrv-doi}

\bibliography{template}
\end{document}